\documentclass{bmcart}

\usepackage{graphicx}
\usepackage{caption}
\usepackage{url}
\usepackage{longtable}
\usepackage{multirow}   
\usepackage{amsthm,amsmath}
\usepackage{amssymb}
\startlocaldefs
\endlocaldefs

\begin{document}

%%% Start of article front matter
\begin{frontmatter}

\begin{fmbox}
\dochead{Research}


\title{A Data-Driven Study of Commonsense Knowledge using the ConceptNet Knowledge Base}

%%%%%%%%%%%%%%%%%%%%%%%%%%%%%%%%%%%%%%%%%%%%%%
%%                                          %%
%% Enter the authors here                   %%
%%                                          %%
%% Specify information, if available,       %%
%% in the form:                             %%
%%   <key>={<id1>,<id2>}                    %%
%%   <key>=                                 %%
%% Comment or delete the keys which are     %%
%% not used. Repeat \author command as much %%
%% as required.                             %%
%%                                          %%
%%%%%%%%%%%%%%%%%%%%%%%%%%%%%%%%%%%%%%%%%%%%%%

\author[
  addressref={aff1},                   % id's of addresses, e.g. {aff1,aff2}
                          % id of corresponding address, if any
%   noteref={n1},                        % id's of article notes, if any
  email={keshen@isi.edu}   % email address
]{\inits{KS}\fnm{Ke} \snm{Shen}}
\author[
  addressref={aff1},
  corref={aff1},
  email={kejriwal@isi.edu}
]{\inits{MK}\fnm{Mayank} \snm{Kejriwal}}

%%%%%%%%%%%%%%%%%%%%%%%%%%%%%%%%%%%%%%%%%%%%%%
%%                                          %%
%% Enter the authors' addresses here        %%
%%                                          %%
%% Repeat \address commands as much as      %%
%% required.                                %%
%%                                          %%
%%%%%%%%%%%%%%%%%%%%%%%%%%%%%%%%%%%%%%%%%%%%%%

\address[id=aff1]{%                           % unique id
  \orgname{Information Science Institute}, % university, etc
  \street{4676 Admiralty Way},                     %
  \postcode{90292}                                % post or zip code
  \city{Marina Del Rey},                              % city
  \cny{US}                                    % country
}
% \address[id=aff2]{%
%   \orgname{Marine Ecology Department, Institute of Marine Sciences Kiel},
%   \street{D\"{u}sternbrooker Weg 20},
%   \postcode{24105}
%   \city{Kiel},
%   \cny{Germany}
% }

%%%%%%%%%%%%%%%%%%%%%%%%%%%%%%%%%%%%%%%%%%%%%%
%%                                          %%
%% Enter short notes here                   %%
%%                                          %%
%% Short notes will be after addresses      %%
%% on first page.                           %%
%%                                          %%
%%%%%%%%%%%%%%%%%%%%%%%%%%%%%%%%%%%%%%%%%%%%%%

\begin{artnotes}
%\note{Sample of title note}     % note to the article
\note[id=n1]{Equal contributor} % note, connected to author
\end{artnotes}

\end{fmbox}% comment this for two column layout

%%%%%%%%%%%%%%%%%%%%%%%%%%%%%%%%%%%%%%%%%%%%%%
%%                                          %%
%% The Abstract begins here                 %%
%%                                          %%
%% Please refer to the Instructions for     %%
%% authors on http://www.biomedcentral.com  %%
%% and include the section headings         %%
%% accordingly for your article type.       %%
%%                                          %%
%%%%%%%%%%%%%%%%%%%%%%%%%%%%%%%%%%%%%%%%%%%%%%

\begin{abstractbox}

\begin{abstract} % abstract
% \parttitle{First part title} %if any

% - what's the problem why it's important(very brief)
% - what we do in the paper(1~2 sentences[connect to title])
% - the conclusion of results. The results show
%- current work belies the lack of commonsense understanding but it still get great process, so the motivation is not such obvious. 
%- in particular .. is not necessary here

Acquiring commonsense knowledge and reasoning is recognized as an important frontier in achieving general Artificial Intelligence (AI). Recent research in the Natural Language Processing (NLP) community has demonstrated significant progress in this problem setting. Despite this progress, which is mainly on multiple-choice question answering tasks in limited settings, there is still a lack of understanding (especially at scale) of the nature of commonsense knowledge itself. In this paper, we propose and conduct a systematic study to enable a deeper understanding of commonsense knowledge by doing an empirical and structural analysis of the ConceptNet knowledge base. 
ConceptNet is a freely available knowledge base containing millions of commonsense assertions presented in natural language. 
%[x; one sentence description of conceptnet that tells us why it's apt for studying commonsense knowledge]. 
Detailed experimental results on three carefully designed research questions, using state-of-the-art unsupervised graph representation learning (`embedding') and clustering techniques, reveal deep substructures in ConceptNet relations, allowing us to make data-driven and computational claims about the meaning of phenomena such as `context'  that are traditionally discussed only in qualitative terms. 
% [x what is the key result? Be specific, without using jargon]. 
Furthermore, our methodology provides a case study in how to use data-science and computational methodologies for understanding the nature of an everyday (yet complex) psychological phenomenon that is an essential feature of human intelligence.

% \parttitle{Second part title} %if any
% Text for this section.
\end{abstract}

%%%%%%%%%%%%%%%%%%%%%%%%%%%%%%%%%%%%%%%%%%%%%%
%%                                          %%
%% The keywords begin here                  %%
%%                                          %%
%% Put each keyword in separate \kwd{}.     %%
%%                                          %%
%%%%%%%%%%%%%%%%%%%%%%%%%%%%%%%%%%%%%%%%%%%%%%


% MSC classifications codes, if any
%\begin{keyword}[class=AMS]
%\kwd[Primary ]{}
%\kwd{}
%\kwd[; secondary ]{}
%\end{keyword}

\end{abstractbox}
%
%\end{fmbox}% uncomment this for twcolumn layout

\end{frontmatter}

\section{Introduction}

Despite the ubiquity of intelligent agents such as Alexa and Siri in modern life, these agents have yet to capture the human element in natural conversations. Even with advances in Natural Language Processing (NLP), deep learning, and knowledge graphs  \cite{Hirschberg261}, \cite{QGuilin}, such agents are merely good at answering questions with a clear structure and `factual need' (e.g.,`What is the weather today in Los Angeles?'). They are still not capable of answering questions with incomplete information or under-specified needs (such as `Should I put my spare change in a piggy bank?') that require more contextual and implicit knowledge that humans often take for granted when navigating daily situations. Among other things, such agents are limited in their \emph{commonsense reasoning} abilities.

Commonsense reasoning is the process that involves processing information about a scenario in the world, and making inferences and decisions by using not only the `explicit' information available to the senses and conscious mind, but also context and implicit information that is based on our `commonsense knowledge'. Commonsense knowledge is difficult to define precisely (not unlike `intelligence') but it may be assumed to be a broad body of knowledge of how the `world' works \cite{Mueller}. Generally, such knowledge is essential for navigating social situations and interactions, physical reasoning (e.g., the simple knowledge that when an object on the table is `picked up', it is not on the table anymore) and (more controversially), knowledge that may rely on local culture and milieu \cite{socialiqa}, \cite{Cyc}. Commonsense knowledge and reasoning have both been recognized as essential for building more advanced `general' AI systems that have human-like capabilities and reasoning ability, even when facing uncertain, implicit (or potentially contradictory) information. Recognizing its importance, researchers in several communities have increasingly engaged in researching and evaluating commonsense reasoning on tasks pertaining to question answering and abductive reasoning \cite{Davis}, \cite{Melissa}, \cite{socialiqa}, \cite{sap-etal-2020-commonsense}. 

\begin{figure}
  \captionsetup{font={small}}
  \centering
  \includegraphics[width=12cm]{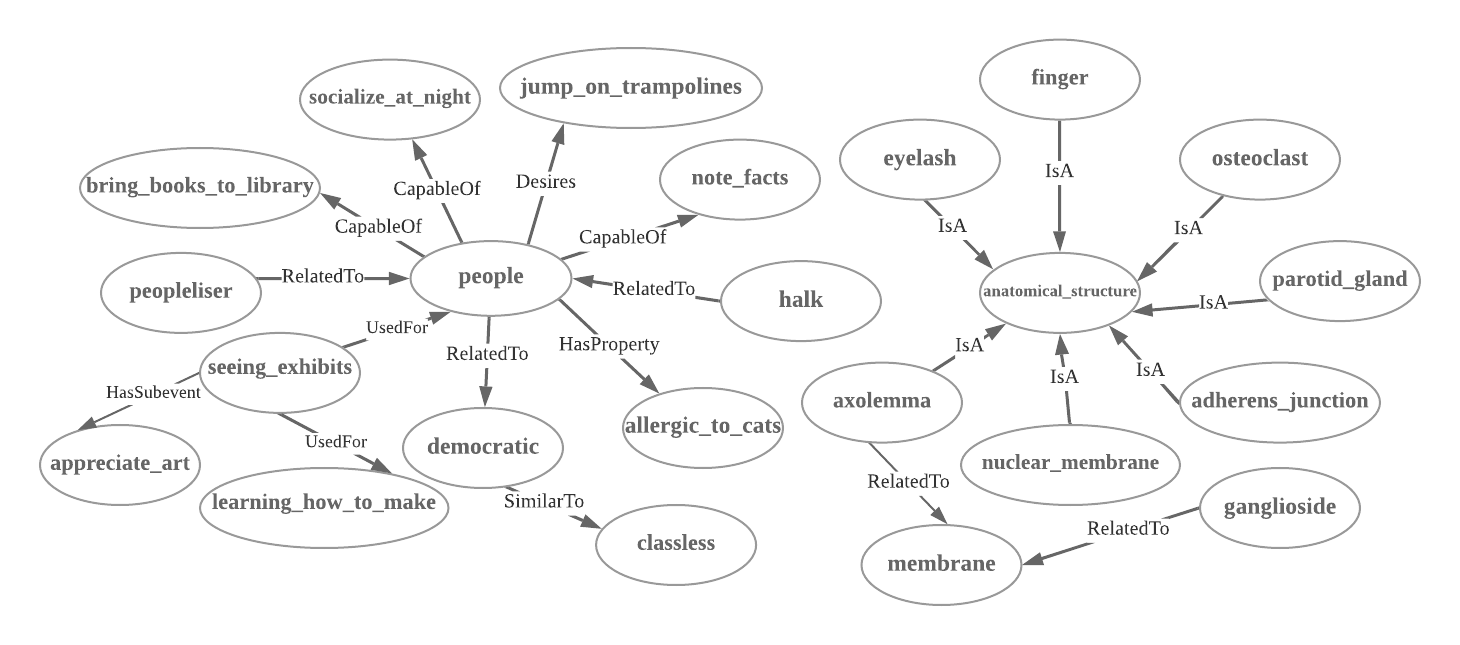}
  \caption{\csentence{A fragment of the ConceptNet Knowledge Graph (KG).}
  Knowledge in ConceptNet is represented as `triples' or 3-tuples, with each triple containing two entities connected by a relation. This fragment contains 21 triples (entities are shown as circles and relations are directed, labeled edges). Unlike certain other KGs (primarily in the Semantic Web community, such as DBpedia), typical versions of ConceptNet do not syntactically distinguish between an entity and the attribute of an entity (e.g., `democratic'). }
  % The triples presented here are selected from those which involved in two main entities `people' and `anatomical\_structure'. Entities are shown in circles. Solid lines with arrowheads show the relations between entities, starting from head entities and ending at tail entities.}
  \label{figure: conNet_frag}
\end{figure}

ConceptNet\footnote{\url{https://conceptnet.io/}.} is a large-scale, freely available knowledge base that describes commonsense knowledge as a set of assertions \cite{Conceptnet5}. It is designed to represent the common knowledge needed to help machines better understand the meaning of words and phrases people use. The \emph{graph structure} that represents knowledge in ConceptNet in patterns of connected word/phrase nodes is particularly useful for textual reasoning over natural language documents. An example of how such knowledge is organized in ConceptNet is illustrated in Figure \ref{figure: conNet_frag}. ConceptNet originated from the Open Mind Common Sense \cite{OMCS} project, itself launched in 1999 at the MIT Media Lab. It is continuously updated to include new knowledge from crowdsourced resources, expert-created resources, and games designed specifically\footnote{Described on the website as `games with a purpose'.} to elicit certain kinds of commonsense annotations, such as knowledge about people's intuitive word associations. 

ConceptNet has some notable differences from other popular knowledge bases. In comparison with WordNet \cite{wordnet}, which focuses on preserving lexicographic information and the relationship between words and their senses, ConceptNet maintains a semantic network structure that is designed to capture commonsense assertions, such as shown in Figure \ref{figure: conNet_frag}. In particular, ConceptNet contains more kinds of relationships than WordNet. Another comparable knowledge base is Cyc \cite{Cyc}, which focuses on standardization of commonsense for efficient logical reasoning. However, ConceptNet is optimized for making inferences over natural language texts, and unlike Cyc, is not proprietary. ConceptNet captures a wide range of commonsense knowledge (such as in Cyc); however, it expresses knowledge in an easy-to-use format (i.e. as sets of 3-tuple assertions, such as illustrated in Figure \ref{figure: conNet_frag}) rather than higher-order logic notation. Because of these advantages, ConceptNet has rapidly emerged as a practical dataset and resource for various kinds of machine learning and NLP in the last ten years in particular \cite{CNapp1}, \cite{CNapp2}, \cite{CNapp3}.

While the utility of ConceptNet as a resource in algorithmic pipelines has been demonstrated, we hypothesize that it can be used as a fundamental resource to understand the nature of commonsense knowledge more precisely. Note that we are not trying to reverse-engineer cognitive processes in the human brain that led to the ingestion and formation of such knowledge, but rather, to yield key findings into some central properties of the knowledge itself. Despite being blended into diverse aspects of daily life, we are not aware of a systematic explanation about the nature of commonsense as a `belief system about the world' that informs most adults' everyday thoughts, behaviors and actions \cite{jwastington}. Researchers have also differed on ideas concerning commonsense knowledge and reasoning \cite{Redekop}, \cite{BSherin}. Therefore, there is strong motivation to gain new insights into commonsense by using a systematic methodology, and easily available and trustworthy data. In this article, we take up this challenge both by presenting a methodology and by showing how the methodology yields useful and novel results when applied to relations and entities in ConceptNet. Specific contributions include:

%The considerable depth and value of knowledge stored in ConceptNet provides an excellent opportunity to achieve a deeper understanding of commonsense. Commonsense is often thought of as a belief system about the world that is generally accepted without debate. This belief system is upheld by most people and informs their everyday thoughts and behaviors as adults \cite{jwastington}. Although commonsense is blended into diverse aspects of daily life, there does not exist a systematic explanation about the nature of commonsense. The characteristics and properties are still unclear to people. Researchers even differed on the ideas about what commonsense is \cite{Redekop}, \cite{BSherin}. To understand commonsense better, this paper presents an in-depth study of ConceptNet in order to obtain new insights about commonsense.
\begin{itemize}
  \item[*] We embed nodes and relations in ConceptNet into a continuous, low-dimension embedding space by a state-of-art graph embedding technique PyTorch-BigGraph(PBG) \cite{pbg} to enable more robust computational analysis in vector space. Using both information theory and standard correlational analysis (as well as a judicious set of visualizations), we show that the embeddings reveal some interesting aspects of ConceptNet relations and their similarities to each other.  Relations in ConceptNet are clearly not independent from one another and share unique degrees of overlap and information content.
  \item[*] For specific high-volume and coarse-grained relations (such as `HasContext\footnote{Defined by ConceptNet as `A is a word used in the context of B, which could be a topic area, technical field, or regional dialect'.}', `SymbolOf\footnote{`A symbolically represents B'.}', and `FormOf\footnote{`A is an inflected form of B; B is the root word of A'.}' that are known to be significant in commonsense reasoning, we discover and explore \emph{substructures} within these relations by using a combination of clustering and qualitative analysis.
  \item[*] We shed light into another important phenomenon (`negation') by contrast the triple structures between an important relation `Desires' and its negation `NotDesires'. In commonsense knowledge, we show that negation does not obey logical intuitions, and that in the embedding space, triples with the negated relation are virtually undistinguishable from triples with the non-negated relation. However, we find that a machine learning model can be trained to recognize these differences. This finding reveals the nature of ``negation'' relation in ConceptNet and the structure properties between a relation and its negation.

\end{itemize}

The rest of this article is structured as follows. In Section 2, we describe relevant lines of research related to this work. In Section 3, we present our experimental study and methodology. Section 4 summarizes and discusses the key findings, and the article concludes in Section 5.

\section{Related Work}\label{relatedwork}

% In the related work, add a section on commonsense reasoning and AI, and another section on knowledge graphs and commonsense. Add yet another section on cognitive science. We want the related work to be more extensive than what you have. You don't have to fill these in yet. Some of the work you have in related work don't belong there and need to move to a later section (I mentioned t-sne, but the description of conceptnet needs to move to a section called 'Data'. If you don't have it in already, make sure to mention relevant stats on conceptnet. Also, make sure to cite Cyc in the knowledge graphs sub-section of related work, if you haven't already
\subsection{Commonsense Reasoning}

Although both Wikipedia and the (relatively) recent DARPA Machine Common Sense (MCS) program\footnote{\url{https://www.darpa.mil/program/machine-common-sense}; \url{https://en.wikipedia.org/wiki/Common_sense}.} define commonsense reasoning as `Äúthe basic ability to perceive, understand, and judge things that are shared by nearly all people and can be reasonably expected of nearly all people without need for debate'Äù, there is no official, sufficiently broad definition that we can cite outside of the psychology context. Within psychology, \cite{smedslund1} has defined commonsense as `the system of implications shared by the competent users of a language'. One important commonality that is shared, no matter the definition or field considered, is that commonsense knowledge is (at least to a degree) \emph{implicit}. It is also not controversial that commonsense knowledge is key to humans being able to interact with the world (and other humans) around them without needing extensive training for every situation, as machines seem to require (e.g., face recognition or information extraction). Despite its seeming simplicity, commonsense reasoning in machines involves several research issues, including but not limited to the representation of given scenario, the representation of commonsense knowledge, context-sensitive effects of events and the representation (and reasoning) of concurrent events \cite{Mueller}. There is very little work on the nature of commonsense knowledge: one rare example is a recent book \cite{csbook1} attempts to provide a `theory' of commonsense by breaking down commonsense knowledge into various categories, and present formalisms for those categories. Other similar work along those lines include \cite{csaxioms1}, \cite{csaxioms2}, \cite{csaxioms3}. Unlike those works, we take an inductive, data-driven approach. Our work is potentially complementary to \cite{csbook1}, since some of our findings may be used in the future to provide evidence for (or against) some of their purely theoretical claims.

Progress has been achieved in specific \emph{kinds} of commonsense reasoning, especially in reasoning about time and internal relations \cite{taxonomy1}, \cite{Pinto}, reasoning about actions and change \cite{Srinivas}, and the sign calculus \cite{Davis}. Semantics have played an important role in some of these successes \cite{semantics1}; in fact, ConceptNet itself has been described as a `semantic network' \cite{conceptnet3}. WordNet \cite{wordnet} and Cyc \cite{Cyc} are two useful resources that allow us, in different ways, to obtain more insight into the semantics of sentences. WordNet is a semantic network whose entries are organized in terms of semantic inter-relationships. The easy-to-use network structure lends it to being frequently applied in multiple reasoning systems \cite{botschen}, \cite{angeli}, \cite{Lin2017}. Cyc focuses on designing a universal schema (a higher-order logic) to represent commonsense assertions, which also supports reasoning systems \cite{Panton06commonsense}, \cite{Ramachandran05} conduct richer logic inferences. 

The increasing enrichment of commonsense knowledge graphs fits with one of the approaches proposed in the aforementioned DARPA MCS program for developing flexible machine commonsense services. Commonsense reasoning systems are typically measured against benchmark tests, some well known examples of such datasets including aNLI \cite{anli}, HellaSwag \cite{hellaswag} and Physical IQA \cite{PIQA}. The other approach strongly supported by DARPA is to develop systems that think and learn as humans do in the very early stages (the first 18 months). Performers in that task aim to develop systems that can provide empirical and theoretical guidance on various aspects of commonsense reasoning abilities in humans by mimicking the core domains of child cognition for objects (intuitive physics), agents (intentional actors), and places (spatial navigation). Our work is more strongly related to the former approach, since ConceptNet primarily comprises assertions of commonsense knowledge that adults, and not necessarily children, would be expected to have. Potentially, our methodology could be applied to other knowledge graphs that are more representative of the knowledge that children have, but we are not aware of any such knowledge graphs currently in existence.

% Nowadays, we construct various knowledge graph to save these knowledges and represent pieces of knowledge as triples. These triples could still reserve their origin language and later be transformed to embedding vectors to manipulate. The key points left are keeping enriching knowledge graph to include more common senses and how to use these knowledge.

\subsection{Knowledge Graphs}

A knowledge graph (KG), of which ConceptNet is an example, is a graph-theoretic way to represent knowledge modeled according to an ontology, typically with the help of subject-matter experts, automatic data extraction, and machine learning algorithms.Rapid advances in KG construction and application have been witnessed in recent years. KGs such as Freebase \cite{freebase}, WordNet \cite{wordnet}, DBpedia \cite{JensDbpedia}, and Yago \cite{yago} have been successfully created and applied to a range of tasks \cite{KGapp1}, \cite{KGapp2}, \cite{KGapp3}. Using these KGs, machine learning and representation leaning algorithms have also advanced with every passing year, in areas as wide ranging as knowledge graph `embeddings', question answering and distant supervision \cite{KGQA1}, \cite{KGDS1}, \cite{KGQA2}, \cite{KGDS2}. Concerning commonsense reasoning, ConceptNet and Cyc \cite{Cyc} (mentioned in the introduction as a proprietary KG that precedes ConceptNet) are the only two examples that we are aware of. Owned by Cycorp Inc., Cyc's knowledge base claims\footnote{\url{https://www.cyc.com/}.} to be the `broadest, deepest, and most complete repository ever developed', but given its proprietary nature, the claim is difficult to validate. It is not known how similar Cyc is to ConceptNet, but potentially, the same methodology proposed in this paper could be applied to Cyc to replicate, strengthen or refute some of our findings.

In contrast, ConceptNet is a freely available multilingual KG that connects everyday entities using a rich set of relations \cite{conceptnet5.5}. As mentioned in the introduction, ConceptNet serves as an important background resource for commonsense reasoning and question answering. However, it has not been studied directly for gaining insights into commonsense knowledge, even though there is precedent. For example, studies on DBpedia and YAGO have been conducted specifically to understand their relational structure and the structural properties of the encyclopedic knowledge that these KGs are known for \cite{DBpediaStudy}, \cite{YAGOStudy}, \cite{KGComp}. We attempt to do the same, but with commonsense knowledge as the focus.  

Another line of work highly related to this one is \emph{representation learning}, especially as they pertain to KGs. Such algorithms take as input the KG, including entities and relations, and embed them into continuous vector spaces, as surveyed by \cite{QWang}, models such as RESCAL \cite{RESCAL}, TransE \cite{transE}, TransH \cite{transh}, KG2E \cite{KG2E}, and RotateE \cite{rotateE} all achieve good performance on various tasks, such as KG completion \cite{transE} and relation extraction \cite{JWeston}, which allows for further improvement of the KG. In this article, we use a graph embedding package that builds on the ones above (especially TransE) and is especially designed for graphs with many millions of triples. 
% \subsection{Cognitive Science}

\section{Empirical Study}

Our guiding principle in this paper is that a commonsense KG such as ConceptNet could be used as the basis for understanding the nature of commonsense knowledge and (potentially) commonsense reasoning. Since this is a broad area of study, we define a specific set of research questions that could be used to shed more light on this subject matter. In this section, we detail critical aspects of the ConceptNet dataset and methodology that are relevant for our empirical investigations. Subsequently, we also investigate each question as separate but related studies, with accompanying descriptions of methodology, results, discussion and implications. Towards the end of the section, we collect these findings and synthesize them into a cohesive set of insights.  

\subsection{Preliminaries and Setup}

Conceptually, ConceptNet may be described as a multi-relational graph $G = (V, R, E)$, where V is the set of entities or \emph{nodes}, R is the set of 49 \emph{relations} and E is the set of triples or \emph{edges} where each triple $e = (h, r, t) \in E \subseteq V \times R \times V$. $h$ and $t$ are respectively referred to as the \emph{source} (or the \emph{head}) and the \emph{destination} (or the \emph{tail}) entity, and $r$ is the relation. For the purposes of maintaining consistent terminology, we use the terms \emph{triple}, \emph{head}, \emph{tail}, and \emph{relation} to refer to the concepts $e$, $h$, $t$ and $r$ respectively. Furthermore, these head and tail entities are collectively referred to as \emph{entities} rather than \emph{nodes}.

Note that $G$ may be thought of in a non graph-theoretic way as a \emph{set} of triples as has been common in the natural language community when discussing these datasets as \emph{knowledge bases} rather than (multi-relational) \emph{knowledge graphs} (where triples are usually interpreted as directed edges \cite{kejriwalDSKG}). The distinction is not relevant for the purposes of this paper, but we do note that it is more natural to think about ConceptNet as a graph due to its structural properties. Earlier, Figure  \ref{figure: conNet_frag} expressed a fragment of ConceptNet as a multi-relational graph (with 21 triples).

Entities and relations are projected into a continuous vector space by PyTorch-BigGraph (PBG), an efficient and state-of-the-art embedding system for large-scale graphs\footnote{In particular, PBG supports multi-entity and multi-relation graph embeddings, unlike network embeddings such as DeepWalk and node2vec \cite{deepwalk}, \cite{node2vec}, and its embedding quality has been found to be comparable with (or even exceed) existing `knowledge graph' embedding systems, evaluated on the Freebase \cite{freebase}, LiveJournal \cite{Livejournal} and YouTube graphs \cite{youtube}.}, for further computation and analysis \cite{pbg}. Next, we describe the raw data and our sampling mechanism for studying the raw data, as well as the setup and training of PBG. 

% \footnote{\url{https://github.com/facebookresearch/PyTorch-BigGraph}}

\subsubsection{Raw Data, Sampling and Representation Learning} 

In our empirical study, we use the latest version of ConceptNet (version 5.7) that was available at the time of writing. 
% \footnote{This version can be downloaded at \url{https://github.com/commonsense/conceptnet5/wiki/Downloads}} 
One important aspect of ConceptNet is the ratio of the number of unique entities to the total number of triples, which is much higher in ConceptNet (0.833) than in other similarly-sized KGs such as Freebase (0.055) or WordNet\footnote{For the interested reader, we are specifically referring to the FB15k and WN18 datasets, as designated in multiple papers on knowledge graph embeddings \cite{SongHJ}, \cite{WZhang}, \cite{FZhao}.} (0.289). Additionally, while ConceptNet contains far more entities than the other KGs, it contains fewer unique relations (the `labels' on the edges, or the size of the set $R$) than the other KGs. 

In practice, these significant deviations in terms of basic entity/relation statistics can cause problems for KG representation learning packages (even those designed for large-scale KGs) including PBG, which is used in this article for experiments. In embedding the full ConceptNet knowledge base, we found that, during the training process, the PBG algorithm fails with a `bus error' message if the number of input triples exceeds 4 million. This seems to be occurring due to the algorithm running out of shared memory (despite the fact that we execute the algorithm on a machine with $61$ GB memory). 
\begin{table}[h!]
\begin{tabular}{|c|c|c|c|c|}

\hline
triple    & entity    & head entity & tail entity & overlap between head and tail entity \\ \hline
4,000,000 & 3,933,840 & 2,781,892    & 1,387,571    & 235,623                                   \\ \hline

\end{tabular}
\caption{The number of triples, entities (including separate head and tail entity counts, and their overlap) in the sample of ConceptNet considered in this study.}
\label{table: triple_statistics}
\end{table}

\begin{table}[!h]
\centering
\resizebox{\textwidth}{45mm}{
\begin{tabular}{|c|c|c|c|c|c|}
         \hline
relation\_name               & triple\_num & entity\_num & relation\_name             & triple\_num & entity\_num \\ \hline
LocatedNear               & 13          & 26         & dbpedia/leader          & 13         & 19          \\ \hline
CreatedBy                 & 14          & 27          & NotHasProperty          & 44          & 81          \\ \hline
NotCapableOf              & 72          & 132         & dbpedia/capital         & 72          & 137         \\ \hline
Entails                   & 73          & 134         & dbpedia/product         & 81          & 140         \\ \hline
dbpedia/knownFor          & 87          & 168         & dbpedia/field           & 114         & 163         \\ \hline
dbpedia/language          & 151         & 181         & dbpedia/occupation      & 183         & 236         \\ \hline
dbpedia/influencedBy      & 210         & 246         & InstanceOf              & 415         & 570         \\ \hline
DefinedAs                 & 433         & 812         & dbpedia/genus           & 464         & 821         \\ \hline
NotUsedFor                & 519         & 833         & HasLastSubevent         & 571         & 867         \\ \hline
dbpedia/genre             & 621         & 759         & ObstructedBy            & 869         & 1,555       \\ \hline
ReceivesAction            & 988         & 1,721       & CausesDesire            & 1,003       & 1,480       \\ \hline
CapableOf                 & 2,146       & 3,348       & MannerOf                & 2,164       & 2,923       \\ \hline
Antonym                   & 2,601       & 5,100       & HasFirstSubevent        & 2,625       & 3,267       \\ \hline
MadeOf                    & 2,715       & 2,936       & HasA                    & 2,962       & 3,908       \\ \hline
HasProperty               & 3,573       & 4,885       & Causes                  & 3,705       & 4,737       \\ \hline
Desires                   & 4,121       & 4,185       & HasPrerequisite         & 4,194       & 4,837       \\ \hline
NotDesires                & 4,263       & 4,186       & AtLocation              & 4,497       & 5,983       \\ \hline
SimilarTo                 & 6,980       & 10,676      & PartOf                  & 7,048       & 9,507       \\ \hline
DistinctFrom              & 10,529      & 16,428      & HasSubevent             & 11,899      & 12,969      \\ \hline
MotivatedByGoal           & 11,996      & 12,186      & UsedFor                 & 13,212      & 15,789      \\ \hline
EtymologicallyDerivedFrom & 46,451      & 78,335      & SymbolOf                & 63,785      & 51,298      \\ \hline
DerivedFrom               & 93,190      & 158,921     & EtymologicallyRelatedTo & 97,124      & 145,853     \\ \hline
IsA                       & 100,451     & 127,922     & HasContext              & 133,035     & 135,211     \\ \hline
FormOf                    & 630,914     & 912,022     & Synonym                 & 1,101,134   & 1,356,240   \\ \hline
\textbf{RelatedTo}                 & \textbf{1,501,359}   & \textbf{1,536,157}   &                            &             &             \\ \hline
\end{tabular}
}
\caption{The numbers of triples and entities corresponding to each of the 49 relations in the ConceptNet sample studied in this article. Note that an entity could occur in multiple triples, each with a different relation. The `RelatedTo' relation, shown in bold, has the most number of triples and entities associated with it.}
    \label{table: statistics of triple and entity}
\end{table}

To address the memory issue and ensure that our results can be extended or replicated in the future using reasonable computation resources, we randomly sampled 4 million triples from ConceptNet for all studies described herein. Key statistics are tabulated in Table \ref{table: triple_statistics}. The head entities-set is twice the size of the tail entities-set and their overlap is approximately 1/20 of the total entities. Cursory analysis also showed that the head entity `/c/en/person' and tail entity `/c/fr/francais' were respectively found to have the most triples associated with them. Other relation-specific statistics are tabulated in Table \ref{table: statistics of triple and entity}. `/r/RelatedTo' was found to be the most frequent relation, occurring in more than 1 million triples.

We input these 4 million triples into the PBG algorithm for representation learning. We partition the 4 million sampled triples into train, validation and test splits, containing 75\%, 12.5\%, 12.5\% of the total triples, respectively.  Before doing the sampling, we also remove triples with the `ExternalURL' relation. \emph{ExternalURL}  is a `non-semantic' relation that is only referring to a URL identifier and cannot be used for structural analysis of the kind proposed in this paper. We conduct training on a single server in the Amazon cloud with 4 Intel Xeon cores (one socket) and 61 GB of RAM. At the end of training, the algorithm outputs a vector for every unique relation and entity in the training dataset.

\subsubsection{Validating Quality of Embeddings}

As mentioned earlier, due to the limits on the memory requirements of PBG, we randomly sampled 4 million triples from the raw ConceptNet knowledge base for representation learning and for the subsequently described studies. Due to the sampling, a reasonable suspicion arises as to whether the quality of the learned representations or embeddings can be trusted. To validate the quality of these embeddings, we designed a series of quantitative measures to analyze the effectiveness of the PBG outputs.

Specifically, having obtained the embeddings output by PBG, we first compute a \emph{centroid vector} for each relation. Recall that we denoted the graph using the symbol $G=(V,R,E)$, where $E$ was the set of triples or `edges' in the graph. In a slight abuse of notation, we use the symbol $G_E$ to represent the set $E$ associated with $G$.  
Given a relation $r \in R$, let $G_r \subseteq G_E$ be the subset of triples in $G_E$ that have relation $r$. For each such triple $(h,r,t)$ in $G_r$, we define the \emph{translation} vector $\vec{v} = \vec{t} - \vec{h}$, where $t$ and $h$ are the embeddings learned by the (previously described) PBG algorithm for entities $t$ and $h$ respectively. The `centroid' of $r$ is then simply the mean of the translation vectors in $G_r$. Formally, the \emph{centroid vector} $\vec{r_c}$ of $r$ is given by the equation below:

\begin{equation}
  \label{eqn:centroid_vector}
  \vec{r_c} = \frac{1}{|G_r|}\sum_{(h,r,t) \in G_r}(\vec{t} - \vec{h})
\end{equation}

Note that this yields two distinct vectors for $r$: the vector directly output by the graph embedding (denoted as $\vec{r}$) and the centroid vector $\vec{r_c}$. We use the symbol $\mathcal{R}$ to indicate the set of direct embeddings for all 49 relations and the symbol $\mathcal{R}_c$ to indicate the set of centroid vectors. The latter is expected to be more robust and interpretable since it is directly derived using translation (between the tail and head entity embeddings, averaged over all the triples where it occurs). 

With this technical machinery in place, we validate our 4 million-triples sample as follows. First, we calculate two similarity lists, $SL_r$ and $SL_r^{'}$, per relation, using each of these two notions of embedding $r$. $SL_r$ denotes the list containing the cosine similarities between $\vec{r}$ and each translation vector in $G_r$ (since in the general case, each triple could yield a different translation vector, even with $r$ fixed), while $SL_r^{'}$ denotes the list of cosine similarities between the centroid vector $\vec{r_c}$ and each translation vector\footnote{Since there is only one $\vec{r_c}$ and relation vector $r$ per relation, the two lists are \emph{aligned} (by virtue of using the same set of translation vectors computed over the triples in $G_r$).}. 

% We then calculate two similarity lists $SL$ for these two vectors by calculating the similarities between $\vec{v_n} \in V$ and $\vec{r} / \vec{r_c}$, $SL_r$ denotes the list for $\vec{r}$ and $SL_r^{'}$ for $\vec{r_c}$.

Given these two per-relation lists and using established statistical and information-theoretic concepts such as Spearman's rank correlation and the Kullback-Leibler (KL) Divergence, we show in the subsequent experiment that the two lists are highly correlated and have low `divergence'. Since $\vec{r_c}$ is a function of the entities in the triples (and never directly uses the relation embedding output by PBG), while $r$ was directly output by PBG, this result shows that the embeddings learned on our sample are highly self-consistent and conform closely to the notion of translation\footnote{In contrast, if $r$ had showed great divergence (or \emph{no} correlation) compared to $r_c$, it would have raised questions about whether the vectors had truly been learned by PBG in a sufficiently non-random way that (at least approximately) resemble the translation operation.} that such neural graph embeddings have been designed to capture in vector space \cite{transE}.
\begin{table}[!h]
    \centering
    \resizebox{\textwidth}{45mm}{
    \begin{tabular}{|c|c|c|c|}
        \hline
\textbf{Relation}                     & \textbf{Spearman's correlation} & \textbf{Relation}                   & \textbf{Spearman's correlation} \\ \hline
IsA                       & -0.773                 & NotDesires              & 0.954                  \\ \hline
dbpedia/knownFor          & 0.795                  & PartOf                  & -0.939                 \\ \hline
HasSubevent               & 0.882                  & dbpedia/genus           & -0.962                 \\ \hline
Entails                   & -0.958                 & EtymologicallyRelatedTo & -0.385                 \\ \hline
DerivedFrom               & -0.864                 & HasA                    & 0.891                  \\ \hline
UsedFor                   & 0.926                  & Desires                 & 0.946                  \\ \hline
CapableOf                 & 0.934                  & dbpedia/leader          & 0.705                  \\ \hline
AtLocation                & 0.600                  & CreatedBy               & 0.780                  \\ \hline
HasContext                & -0.516                 & NotUsedFor              & 0.639                  \\ \hline
Antonym                   & -0.856                 & DefinedAs               & 0.812                  \\ \hline
HasLastSubevent           & 0.918                  & SymbolOf                & 0.861                  \\ \hline
CausesDesire              & -0.946                 & LocatedNear             & -0.951                 \\ \hline
EtymologicallyDerivedFrom & -0.865                 & HasPrerequisite         & 0.797                  \\ \hline
InstanceOf                & -0.947                 & MadeOf                  & 0.921                  \\ \hline
dbpedia/influencedBy      & -0.475                 & ReceivesAction          & 0.979                  \\ \hline
MannerOf                  & -0.979                 & dbpedia/capital         & 0.946                  \\ \hline
dbpedia/language          & -0.595                 & Causes                  & 0.987                  \\ \hline
HasProperty               & 0.924                  & NotHasProperty          & -0.736                 \\ \hline
dbpedia/product           & -0.880                 & NotCapableOf            & -0.598                 \\ \hline
HasFirstSubevent          & 0.818                  & dbpedia/field           & -0.611                 \\ \hline
dbpedia/genre             & -0.983                 & SimilarTo               & -0.918                 \\ \hline
DistinctFrom              & 0.756                  & MotivatedByGoal         & 0.957                  \\ \hline
dbpedia/occupation        & -0.591                 & ObstructedBy            & 0.849                  \\ \hline
FormOf                    & -0.708                 & RelatedTo               & -0.937                 \\ \hline
Synonym                   & -0.738                 &                            &                        \\ \hline
    \end{tabular}
    }
    \caption{The Spearman's rank correlation score between $SL_r$ and $SL_r^{'}$. The methodology for constructing these two (aligned) lists is described in the text.}
\label{table: speamans's score}
\end{table}
%\newpage

Spearman's rank correlation was designed to measure both the strength and direction of association between two ranked variables and ranges from -1 (perfect negative correlation) to 1 (perfect positive correlation). Because of the geometric features of the embedding space, we are interested in \emph{whether} there is correlation (i.e. the strength), rather than the direction of the correlation. For this reason, given the two aligned lists per relation ($SL_r$ and $SL_r^{'}$), we computed the \emph{absolute value} of the Spearman's rank correlation for each relation in Table \ref{table: speamans's score} \footnote{As expected, some of the correlations were indeed negative. Of the 49 relations, 24 relations have a Spearman's rank correlation greater than 0.6; however, 25 other relations have negative correlations. In no case is the absolute value less than 0.4.}. 
The consistency of the embeddings are further supported by the information-theoretic KL-Divergence measure, detailed in Appendix A. 

\subsection{Study 1: Relational Structure of ConceptNet}

A key contribution of our study is to classify and study the \emph{relational structure} of ConceptNet by using graph embeddings. Our goal is to use such a study to understand the relational structure of commonsense knowledge itself, since ConceptNet primarily contains commonsense facts and relations.

Given that ConceptNet is a large, crowdsourced commonsense knowledge base with semantically diffuse\footnote{By semantically diffuse, we intuitively mean that \emph{some} relations (e.g., FormOf) are very broad, and relations may also \emph{overlap} with each other in terms of usage, intent or semantics.} relations, a specific question that arises is: how \emph{similar} are two relations? Since there are multiple ways to quantify or even define `similarity', we explore three broad empirical methodologies for answering this question. While one of the methodologies is based on the graph embeddings that we learned earlier, two others are more conservative and traditional. While there is some overlap across the findings yielded by applying all three methodologies, there are also clear differences. In fact, even when using the embedding-based methodology, we show that the choice between using embeddings directly output by PBG and the `centroid' relation embeddings computed using Equation \ref{eqn:centroid_vector}, can yield nuanced results. It is important to note that a methodology should not be thought of (in the context of defining and quantifying `similarity') as `correct' or `incorrect'; rather, our goal is to show the different aspects of relation similarity that the methodologies can collectively reveal.  Next, we detail the three methodologies and the corresponding findings.

\subsubsection{Using Explicit Relation Definitions}
\label{relation_def}

Perhaps the most obvious way of comparing relations to one another is to do a `textual' comparison of their explicitly declared definitions. Taking relations HasContext and PartOf as examples, HasContext is defined as ``A is a word used in the context of B\footnote{ConceptNet relation definitions can be found in \url{https://github.com/commonsense/conceptnet5/wiki/Relations}. Additionally, we reproduce them in Appendix B.}'' and PartOf is defined as ``A is a part of B''. We use a traditional bag-of-words technique, the TF/IDF embedding with cosine similarity, to compare two definitions. With the `vocabulary' consisting of all words that occur in all definitions of relations in ConceptNet, we compute the TF/IDF vector by treating each definition as a `document', followed by the calculation of the cosine similarity between two vectors (corresponding to two definitions) to derive a conservative measure of the similarity between two relations. For the example above (HasContext and PartOf), the cosine similarity between the TF/IDF vectors of HasContext and PartOf is 0.178. 

For each relation, we can compute the relation to which it has the strongest similarity using the methodology above. The results are tabulated for 33 \emph{main}\footnote{On the ConceptNet5 website in the previous footnote, these so-called `main' relations are the ones with definitions.} relations in Table \ref{table: most_similar_relations_def}. We find that the closest relation of HasContext is UsedFor, even though the definition of UsedFor states that ``A is used for B", which is different semantically than saying that ``A is a word used in the context of B" (though the two are not contradictory). This simple procedure illustrates why it is important to use the underlying data in the knowledge base to truly understand the empirical connections between relations; merely considering the surface form of definitions yields a very incomplete (if not misleading) picture. While word embeddings and other sophisticated methods could address some of the limitations of TF/IDF (including synonyms), the basic problem still remains unresolved i.e., explicit comparison of relational definitions tells us nothing about the way in which they have been asserted in triples, in practice.  
\begin{table}[!h]
    \centering
    % \resizebox{\textwidth}{55mm}{
    \begin{tabular}{|c|c|c|}
    \hline
        \textbf{Relation}                      & \textbf{Closest relation}          & \textbf{Score} \\ \hline
RelatedTo                  & ObstructedBy            & 0.226 \\ \hline
FormOf                     & UsedFor                 & 0.467 \\ \hline
lsA                        & PartOf                  & 0.390 \\ \hline
PartOf                     & MadeOf                  & 0.503 \\ \hline
HasA                       & HasPrerequisite         & 0.194 \\ \hline
UsedFor                    & FormOf                  & 0.467 \\ \hline
CapableOf                  & DistinctFrom            & 0.300 \\ \hline
AtLocation                 & UsedFor                 & 0.463 \\ \hline
Causes                     & HasSubevent             & 0.380 \\ \hline
HasSubevent                & Causes                  & 0.381 \\ \hline
HasFirstSubevent           & HasLastSubevent         & 0.752 \\ \hline
HasLastSubevent            & HasFirstSubevent        & 0.752 \\ \hline
HasPrerequisite            & SimilarTo               & 0.302 \\ \hline
HasProperty                & ReceivesAction          & 0.206 \\ \hline
MotivatedByGoal            & CausesDesire            & 0.215 \\ \hline
ObstructedBy               & FormOf                  & 0.380 \\ \hline
Desires                    & UsedFor                 & 0.261 \\ \hline
CreatedBy                  & IsA                     & 0.247 \\ \hline
Synonym                    & EtymologicallyRelatedTo & 0.291 \\ \hline
Antonym                    & HasSubevent             & 0.216 \\ \hline
DistinctFrom               & PartOf                  & 0.330 \\ \hline
DerivedFrom                & RelatedTo               & 0.000 \\ \hline
SymbolOf                   & HasSubevent             & 0.284 \\ \hline
DefinedAs                  & HasProperty             & 0.206 \\ \hline
ReceivesAction             & HasProperty             & 0.206 \\ \hline
MannerOf                   & SimilarTo               & 0.356 \\ \hline
LocatedNear                & HasSubevent             & 0.226 \\ \hline
HasContext                 & UsedFor                 & 0.277 \\ \hline
SimilarTo                  & MannerOf                & 0.356 \\ \hline
EtymologicallyRelatedTo    & Synonym                 & 0.291 \\ \hline
EtymologicallyDerivedFrom  & UsedFor                 & 0.241 \\ \hline
CausesDesire               & Motivated ByGoal        & 0.215 \\ \hline
MadeOf                     & PartOf                  & 0.503 \\ \hline
    \end{tabular}
    % }
    \caption{The most similar relation for each relation, obtained using the TF/IDF methodology described in Section \ref{relation_def}.}
\label{table: most_similar_relations_def}
\end{table}

%On the other hand, similar semantic meaning could be presented in different expressions such as using synonyms and various tense. One example should be highlighted here is SymbolOf. Its official definition is A symbolically represents B. From semantic understanding, it should be similar to DefinedAs, which explained as A and B overlap considerably in meaning, and B is a more explanatory version of A. However, except subject A and object B, there does not exist any word overlap in these two definitions. These reasons will reduce the credibility and reliability of TF/IDF result, at least manual check are needed after the calculation.

\subsubsection{Using Overlap of Head and Tail Entities}
\label{triple_overlap}

The previous method is restricted by the combination of words chosen to express the relation's definition (which may not capture actual usage or `practical' semantics of the relations). In order to capture usage-based semantics of relations (and their corresponding similarity to one another), the triples that assert that relation, rather than the surface form of the relation definition, can be expected to describe the relation more comprehensively. An alternative approach then is to evaluate the similarity between two relations by (intuitively) measuring the overlap between their participating entity sets. Specifically, given the graph $G_E$ (represented as a set of triples, as noted earlier) and two relations $r_1$ and $r_2$ between which we wish to compute the similarity $sim(r_1,r_2)$, let $H_{r_1}$ be defined as the set of all head entities such that the head entity occurs in some triple in $G_E$ with relation $r_1$. Similarly, $H_{r_2}$ can be defined. 
\begin{table}[!h]
    \centering
    \begin{tabular}{|c|c|c|}
        \hline
        \textbf{Relation}                      & \textbf{Closest relation}             & \textbf{Score} \\ \hline
RelatedTo                  & Synonym                   & 0.057 \\ \hline
FormOf                     & RelatedTo                 & 0.008 \\ \hline
IsA                        & Synonym                   & 0.016 \\ \hline
PartOf                     & UsedFor                   & 0.038 \\ \hline
HasA                       & MadeOf                    & 0.076 \\ \hline
UsedFor                    & AtLocation                & 0.112 \\ \hline
CapableOf                  & NotDesires                & 0.095 \\ \hline
AtLocation                 & UsedFor                   & 0.112 \\ \hline
Causes                     & HasSubevent               & 0.122 \\ \hline
HasSubevent                & HasPrerequisite           & 0.163 \\ \hline
HasFirstSubevent           & HasPrerequisite           & 0.148 \\ \hline
HasLastSubevent            & HasPrerequisite           & 0.215 \\ \hline
HasPrerequisite            & HasLastSubevent           & 0.215 \\ \hline
HasProperty                & HasA                      & 0.058 \\ \hline
MotivatedByGoal            & HasFirstSubevent          & 0.105 \\ \hline
ObstructedBy               & DefinedAs                 & 0.063 \\ \hline
Desires                    & NotDesires                & 0.161 \\ \hline
CreatedBy                  & NotHasProperty            & 0.018 \\ \hline
Synonym                    & RelatedTo                 & 0.057 \\ \hline
Antonym                    & RelatedTo                 & 0.057 \\ \hline
DistinctFrom               & IsA                       & 0.013 \\ \hline
DerivedFrom                & NotDesires                & 0.034 \\ \hline
SymbolOf                   & ObstructedBy              & 0.063 \\ \hline
DefinedAs                  & HasProperty               & 0.03  \\ \hline
ReceivesAction             & HasProperty               & 0.03  \\ \hline
MannerOf                   & Entails                   & 0.008 \\ \hline
LocatedNear                & ReceivesAction            & 0.002 \\ \hline
HasContext                 & RelatedTo                 & 0.038 \\ \hline
SimilarTo                  & HasContext                & 0.003 \\ \hline
EtymologicallyRelatedTo    & EtymologicallyDerivedFrom & 0.049 \\ \hline
EtymologicallyDerivedFrom & EtymologicallyRelatedTo   & 0.049 \\ \hline
CausesDesire               & Causes                    & 0.065 \\ \hline
MadeOf                     & HasA                      & 0.076 \\ \hline
    \end{tabular}
    \caption{The most similar relation corresponding to each relation, obtained by choosing the one that has the highest Jaccard similarity when comparing the head entity sets of the two relations.}
\label{table: most_similar_relations_head}
\end{table}

Given the two sets $H_{r_1}$ and $H_{r_2}$ above, we compute $sim(r_1,r_2)$ as the Jaccard similarity\footnote{Defined as $\frac{H_{r_1} \cap H_{r_2}}{H_{r_1} \cup H_{r_2}}$} between $H_{r_1}$ and $H_{r_2}$. We tabulate the results in Table \ref{table: most_similar_relations_head}, wherein, for each relation, we note the relation to which it is most similar, along with the Jaccard similarity score as defined above. We conduct a symmetric exercise, but using the tail entities instead of the head entities (tabulated in Table \ref{table: most_similar_relations_tail} in the Appendix). Taken together, the two tables yield the following insights:

\begin{enumerate}
    \item We find that a few relations have the same most similar relations regardless of whether we choose the closest relation according to their head-entities overlap, or using the TF/IDF methodology in the previous section. For example, AtLocation has  UsedFor as its most similar relation in both Tables \ref{table: most_similar_relations_head} and \ref{table: most_similar_relations_def}, and ReceivesAction is most similar to HasProperty Tables \ref{table: most_similar_relations_head} and \ref{table: most_similar_relations_def}. Therefore, it is not the case that the two methodologies are necessarily in conflict with one another; sometimes, usage and surface semantics coincide. In most cases, however, they do not. Taking again the example of HasContext, we find that in Table \ref{table: most_similar_relations_head} it is most similar to RelatedTo, as we would intuitively expect, while in Table \ref{table: most_similar_relations_def} (as explained earlier) it is most similar to UsedFor, with which is has higher surface (but not necessarily semantic) overlap.
    \item  When we compare the differences between head-entities and tail-entities overlap methodologies, we find that, while there is some agreement on relations such as FormOf and Synonym, there are differences in other relations such as SimilarTo and MadeOf. In looking at the varying results between Tables \ref{table: most_similar_relations_head} and \ref{table: most_similar_relations_tail}, we find that the former is more accurate and makes the most sense. For example, according to the former, the closest relation to MadeOf is HasA, while the latter declares it to be Desires, which is not intuitively plausible. Also, the frequent occurrence of RelatedTo in Table \ref{table: most_similar_relations_head} is a noteworthy example of a `jack of all trades' relation that is the closest relation to several more specific relations. One reason is that RelatedTo is used in very diverse contexts, and this diversity leads to a larger overlap between entity sets. Methodologically, it may be apt to treat these more general and diverse relations differently from more specific ones (such as LocatedNear)
    % [x next, talk about how the tail/head compares ]
\end{enumerate}

\subsubsection{Using Graph Embeddings}
\label{embedding}
While the previous methodologies for measuring relation similarity had utility, their robustness was clearly limited. Graph embeddings are an efficient means to transform nodes (entities) and edges (relations) in the knowledge graph into vectors. Those vectors maximally preserve information and properties of entities and relations, especially the `structural properties' of information--entities or relations close in the graph are mapped to vectors that are also close (in the vector space). We claim that graph embeddings provide us with an effective empirical construct for quantifying the similarity of two relations while addressing robustness issues. 

Earlier, we presented our methodology for obtaining the embeddings from a sampled set of ConceptNet assertions using the PBG algorithm. We also showed that there are two ways to get a valid vector for every relation in $R$ (the set of relations in ConceptNet). One option is to use the `direct' relation vectors as output by the PBG algorithm, but we are also interested in interpreting the relation embeddings in the context of the triples that the relation occurs in. To this end, we also employ $\vec{r_c}$ (the centroid vector defined in Equation 1) in the analysis below. 

One way to analyze the embeddings is to visualize them using dimensionality reduction techniques such as t-Distributed Stochastic Neighbor Embedding (t-SNE) algorithm \cite{t-sne}, followed by visual analysis and inspection. While this has some benefits, including obtaining an `intuitive' feel of the embeddings by seeing their dispersion in a 2D space, visualization methods such as t-SNE do not allow us to formally quantify the similarities between relations in a systematic way\footnote{Nevertheless, we do provide auxiliary analyses using t-SNE visualizations in Appendix C. Therein, we also show that, while t-SNE does reveal some interesting aspects of the relations, it is unable to avoid the issue of information loss compared to cosine-similarity heatmaps.}. Additionally, dimensionality reduction can lose information due to projection in lower dimensions. Therefore, we chose to quantify inter-relation similarities by computing the cosine similarity between each pair of centroid embeddings and (separately), direct embeddings, and plotting them on a square (and symmetric) heatmap. Figures \ref{figure: centroid_heatmap} and \ref{figure: relation_heatmap} respectively illustrate these results for the two different embedding methodologies. The heatmaps reveal some intriguing, though intuitively plausible, insights that we summarize below:
%From the heatmaps we see that SimilarTo and Synonym are close both in Figures \ref{figure: centroid_heatmap} and \ref{figure: centroid_distribution}, as are MadeOf and NotDesires.

\begin{figure}[h!]
  \captionsetup{font={small}}
  \centering
  \includegraphics[width=12cm]{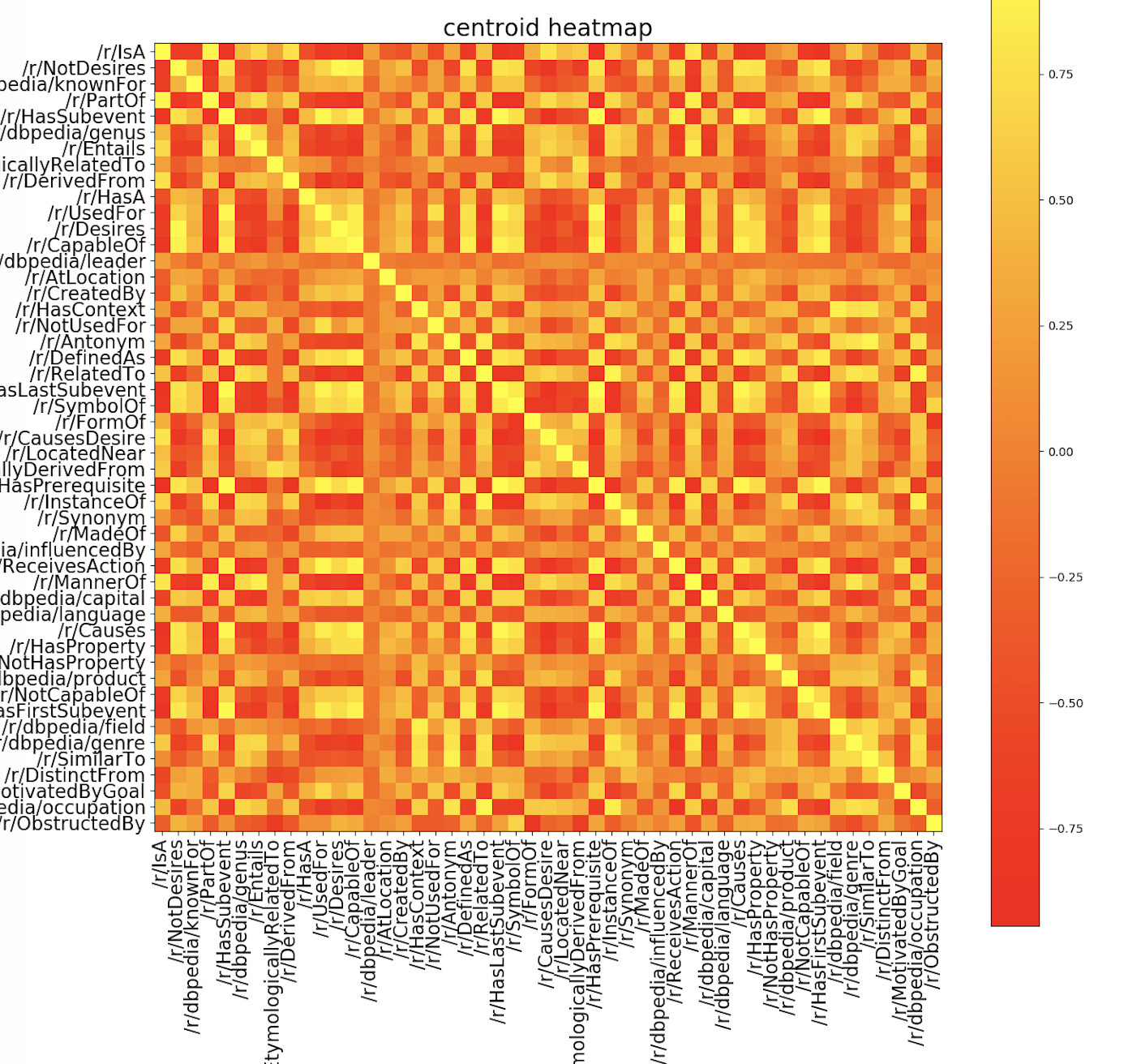}
  \caption{\csentence{The heatmap of cosine similarities between 49 relations' centroid vector}
      The centroid vector for each relation has been described in the text. In the heatmap, a redder shade implies that the corresponding relation pair's centroid vectors' cosine similarity is low (tending to $-1.0$) while yellower shades represent high cosine similarities. The cells in the main diagonal represent the cosine similarity between a relation's centroid vector to itself, which are all shown in yellow, since they equal 1.0 by definition.
      % The heatmap that describes similarities between relation centroid vectors.
      }
  \label{figure: centroid_heatmap}
\end{figure}

\begin{figure}[h!]
  \captionsetup{font={small}}
  \centering
  \includegraphics[width=12cm]{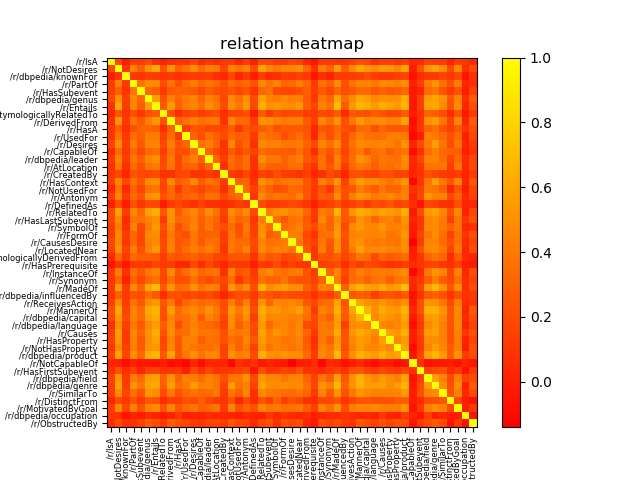}
  \caption{\csentence{The heatmap of cosine similarities between 49 relations' relation embedding}
      The relation embeddings are directly yielded by PBG. The methodology is consistent with that of Figure \ref{figure: centroid_heatmap}.}
  \label{figure: relation_heatmap}
\end{figure}

\begin{enumerate}
    \item Some relations have high similarity with their negations. For example, Desires and NotDesires, UsedFor and NotUsedFor, and Synonym and Antonym. Their closeness can be explained by a number of reasons, including high overlap in their participating entity-sets. We will provide a detailed analysis about a relation example and its negation in Study 3.
    \item A relation generally possesses a high similarity to its `inclusion' relation. An inclusion relation $r_2$ of relation $r_1$ implies that the triples-set in $r_2$ is expected (to a large degree) to be a subset of the triples-set of $r_1$, according to relation definitions. For example, HasFirstSubEvent and HasLastSubEvent are inclusion relations of HasSubEvent, and CausesDesire is an inclusion relation of Causes. We find that such relations show high similarities with their entailed relation in the heatmap, even though an auxiliary analysis using t-SNE visualizations suggest otherwise (Appendix C).
\end{enumerate}

It is also instructive to understand these results in the context of the two previous methodologies described in Sections \ref{relation_def} and \ref{triple_overlap}. In some respects, the results are not very different qualitatively. An example is that, as demonstrated in the heatmap,  relations tend to be close to their inclusion relations due to the overlap between participating entity sets, regardless of whether we use overlap directly (as in the previous section) or a graph embedding methodology as in this section. In Table \ref{table: most_similar_relations_tail}, for example, HasSubEvent was inferred as the closest relation to HasLastSubEvent.

\subsection{Study 2: Substructure detection}
\label{substructure}
Certain relations in ConceptNet are deliberately designed to be broad. A good example is the \emph{HasContext} relation, which is defined on the ConceptNet website as: \emph{A HasContext B} is declared in the knowledge base if `A is a word used in the context of B, which could be a topic area, technical field, or regional dialect'.

We hypothesize that, despite being originally defined so broadly, there is considerable \emph{substructure} in these relations. In considering the definition of HasContext above, multiple contexts are suggested (e.g., \emph{technical field, regional dialect}, and presumably, other contexts that may be similar to these explicit cases). Another example is a relation such as FormOf, where a triple \emph{A FormOf B} may be declared if `A is an inflected form of B; B is the root word of A'. Even the basic official definition \emph{suggests} breadth, since A could either be an `inflected' form of B, or the `root word' of B. Furthermore, there is nothing in the definition that places a strict constraint on such triples, either in theory or in practice. Since ConceptNet is crowdsourced (at least to a great extent), it is quite likely that many people have interpreted these relations at `face value' i.e. in accordance with what one would understand their `everyday' meaning to be. Therefore, the set of experiments described in this section is to measure and describe the \emph{empirical substructure}, if any, of three specific relations (\emph{HasContext, FormOf} and \emph{SymbolOf}) using a systematic methodology. The \emph{SymbolOf} relation is succinctly defined as: the triple \emph{A SymbolOf B} is asserted in the knowledge base if `A symbolically represents B'.  An important aspect of these three relations is not just that they are defined broadly (i.e. are \emph{coarse-grained}) but are also (relatively) \emph{high-volume}. Within our sample of 4 million triples, HasContext, FormOf and SymbolOf are asserted in 133,038, 630,914 and 63,785 triples respectively. This provides an added incentive to study these relations further, since they are clearly central to the knowledge base and its purpose of capturing commonsense knowledge as sets of assertions. 

An established \emph{unsupervised} methodology for discovering structure in large collections of data points is \emph{clustering} \cite{clustering}. The relations in ConceptNet were meant to capture common, informative patterns from various data sources that fed into ConceptNet, as well as crowdsourcing. If well-defined clusters exist, it proves that these coarse-grained relations are not as coarse-grained as the volume of triples suggests. By studying the clusters (including the consistency and subjective nature of data within each cluster, as well as the approximate number of `optimal' clusters revealed through application of an algorithm such as k-Means), we can start gaining insight into each relation. These insights allow us to gain an empirical understanding of words such as `context' and `form' that are important in commonsense reasoning and communication, beyond their formally stated (and rather broad) definitions.   

Before diving into the results for each of these relations, the question of methodology still remains. Most established clustering algorithms require the collection and representation of data points to be described in advance. In this case, since we are interested in studying \emph{each} independent assertion (and thereby, use) of (each of) the three relations, we would ideally like to have a data point for each triple where the relation of interest occurs. Our treatment in earlier sections already provided a mechanism for doing so i.e. obtaining a data point for each relation $r$ per triple $(h,r,t)$ where that relation is asserted by using the translation vector $\vec{t}-\vec{h}$, and where the entity embeddings $\vec{t}$ and $\vec{h}$ are obtained using the PBG package, just as in Study 1. Since HasContext, FormOf and SymbolOf are respectively asserted in 133,038, 630,914 and 63,785 triples within our sample, we obtain the same (respective) numbers of data points (as translation vectors) for each of the three relations. Furthermore, because of the large numbers of data points and the lack of a task-specific objective function (or alternatively, training labels), as well as the preference for using an established and reasonably robust clustering algorithm, we chose to use the classic k-Means algorithm \cite{k-means}. Briefly, k-Means works iteratively to \emph{partition} the dataset into k clusters, each of which is disjoint (since the clusters constitute a partition).

Note that $k$ is a hyperparameter that must be predefined prior to executing the algorithm. There are several ways to obtain the `best' value of $k$ given a collection of points. The underlying commonality between these methods is to compute a `score' (typically an error score, implying that lower is better) for each value of $k$ (over some predetermined range and increment over the range) after executing k-means for that $k$. Next, we plot the score vs. $k$ and look for sudden shifts in the \emph{second derivative} of the curve\footnote{Although the curve can be monotonic for some methods, it is not always guaranteed. Hence, it is incorrect to look for a `minimum'.}. Intuitively, this looks like a clear `bend' in the curve.  

One such candidate method is the \emph{elbow method}, which computes an error-based score based on the `dispersion' of points within each cluster. However, as the first subplot in Figure \ref{figure:hascontext-para} shows\footnote{In Appendix D, we reproduce analogous subplots (Figures \ref{figure:symbolof-para} and \ref{figure:formof-para}) for the SymbolOf and FormOf relations.}, there is no visible decline in the second derivative with $k$ (i.e. the bend, if it exists, is far too subtle to be useful). In considering three alternative established methods--namely, the Silhouette Coefficient method, the Davies-Bouldin Index, and the Calinski-Harabasz Index\footnote{The Silhouette Coefficient value measures how similar a point is to its own cluster's centroid (cohesion) compared to other clusters' centroids (separation). The Index measures compute their scores in slightly different ways, but with the same underlying philosophy that clusters should be cohesive and well-separated. Specific details and formulae may be found in the cited works.} \cite{SC}, \cite{cluster_measure}, we find that $k=20$ is a robust choice for all the three relations.     

\begin{figure}[htbp]
  \begin{minipage}[t]{0.31\textwidth}
    \centering
    \includegraphics[width=1.7in, height=2.0in]{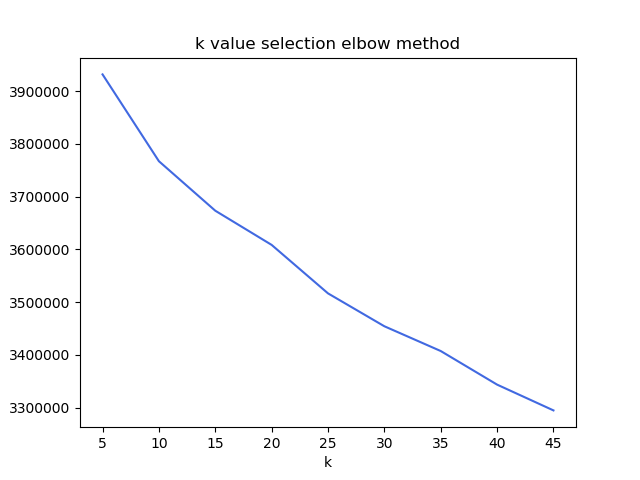}
    % \caption{HasContext\_EM}
  \end{minipage}
  \begin{minipage}[t]{0.31\textwidth}
    \centering
    \includegraphics[width=1.7in, height=2.0in]{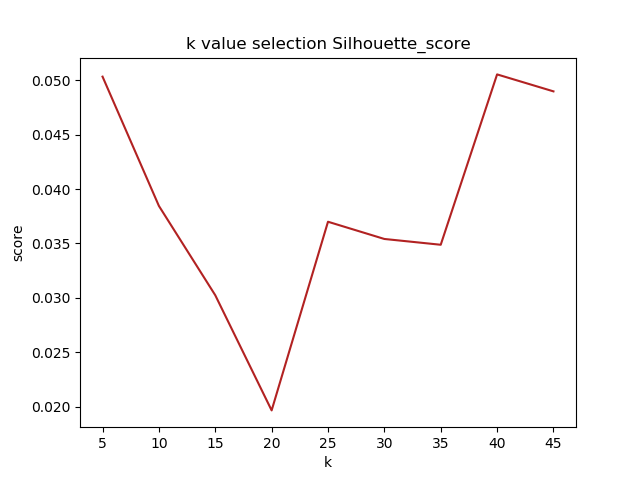}
%   \caption{HasContext\_Sil}
  \end{minipage}
  \begin{minipage}[t]{0.31\textwidth}
    \centering
    \includegraphics[width=1.7in, height=2.0in]{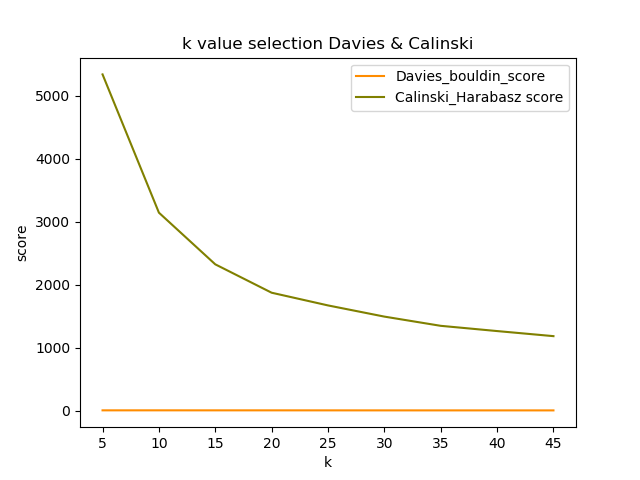}
% \caption{HasContext\_DC}
  \end{minipage}
  \caption{\csentence{Selection of the parameter k for k-means clustering of HasContext triples.} The line chart describes the trend in the HasContext triples' clustering scores with different values of k. The clustering score used in the subplots (from left to right), is measured by the elbow method, Silhouette Coefficient, Davies-Bouldin index, and Calinski-Harabasz index, respectively (with the last two in the same subplot). The x-axis shows the k value and the score is shown on the y-axis.}
    \label{figure:hascontext-para}
\end{figure}

With this value of $k$ ($=20$) in place, we conduct k-Means clustering for each of the three relations, and perform dimensionality reduction using t-SNE to obtain a visualization of all the clusters and triples\footnote{Unlike the previous study, a heatmap-style study is not possible here, due to the large numbers of triples and data points involved. Nevertheless, we do quantify the clusters using separation and cohesion measures in Tables \ref{table: average_separation} and \ref{table: average_cohesion}.}. In describing the discovered substructures (`clusters') for each relation below, we define and quantitatively analyze the \emph{cohesion} and \emph{separation} of these sub-clusters in Table \ref{table: average_cohesion} and \ref{table: average_separation}; we also sample some triples and study them qualitatively in Appendix D to comment on the nature of triples within the substructures and provide some subjective validation of the k-Means output.  

\begin{table}[h!]
\begin{tabular}{|r|r|r|r|}
\hline
\multicolumn{1}{|l|}{\textbf{Cluster id}} & \multicolumn{1}{l|}{\textbf{FormOf}} & \multicolumn{1}{l|}{\textbf{HasContext}} & \multicolumn{1}{l|}{\textbf{SymbolOf}} \\ \hline
0                      & 4.517                       & 5.953                           & 5.358                         \\ \hline
1                      & 4.576                       & 5.386                           & 4.815                         \\ \hline
2                      & 4.085                       & 4.661                           & 5.436                         \\ \hline
3                      & 4.254                       & 4.640                           & 5.578                         \\ \hline
4                      & 4.785                       & 5.738                           & 4.703                         \\ \hline
5                      & 4.503                       & 5.555                           & 4.735                         \\ \hline
6                      & 4.660                       & 3.284                           & 3.677                         \\ \hline
7                      & 3.601                       & 5.072                           & 4.854                         \\ \hline
8                      & 3.966                       & 4.051                           & 4.623                         \\ \hline
9                      & 4.547                       & 3.731                           & 4.730                         \\ \hline
10                     & 4.065                       & 4.297                           & 6.319                         \\ \hline
11                     & 4.741                       & 3.371                           & 4.553                         \\ \hline
12                     & 4.214                       & 4.215                           & 4.647                         \\ \hline
13                     & 3.918                       & 5.398                           & 4.031                         \\ \hline
14                     & 5.300                       & 4.331                           & 4.715                         \\ \hline
15                     & 4.451                       & 3.739                           & 4.585                         \\ \hline
16                     & 4.648                       & 5.515                           & 4.443                         \\ \hline
17                     & 4.684                       & 4.987                           & 5.132                         \\ \hline
18                     & 5.151                       & 3.677                           & 4.533                         \\ \hline
19                     & 4.119                       & 3.649                           & 5.149                         \\ \hline
{\bf Mean}             & 4.439                       & 4.562                           & 4.831                         \\ \hline
{Std. Dev.}            & 3.302                       & 13.588                          & 6.084                        \\ \hline
\end{tabular}
\caption{The average cohesion of FormOf, HasContext and SymbolOf clusters, along with their mean and standard deviation. Cohesion is defined in the text (Study 2). Note that cluster IDs are assigned arbitrarily and independently across relations, and not `aligned' in any way. The IDs are re-used again in Figures \ref{figure: hascontext}, \ref{figure: FormOf}, \ref{figure: SymbolOf}, and \ref{figure: SymbolOf_example}, as well as in Tables \ref{table: HasContext} and \ref{table: FormOf} to report on further results.}
\label{table: average_cohesion}
\end{table}

\begin{table}[h!]
\begin{tabular}{|r|r|r|r|}
\hline
\textbf{Cluster id} & \textbf{FormOf} & \textbf{HasContext} & \textbf{SymbolOf} \\ \hline
0           & 2.563  & 4.644      & 2.421    \\ \hline
1           & 2.911  & 3.624      & 2.597    \\ \hline
2           & 2.969  & 3.943      & 5.611    \\ \hline
3           & 2.866  & 5.811      & 2.661    \\ \hline
4           & 2.681  & 4.044      & 2.750    \\ \hline
5           & 2.686  & 3.943      & 2.607    \\ \hline
6           & 3.376  & 6.016      & 4.862    \\ \hline
7           & 3.105  & 3.564      & 2.454    \\ \hline
8           & 2.952  & 4.345      & 2.245    \\ \hline
9           & 2.658  & 5.738      & 2.348    \\ \hline
10          & 4.092  & 4.323      & 4.151    \\ \hline
11          & 2.536  & 6.412      & 2.585    \\ \hline
12          & 2.831  & 4.620      & 2.381    \\ \hline
13          & 3.235  & 3.727      & 4.697    \\ \hline
14          & 4.055  & 4.889      & 2.269    \\ \hline
15          & 2.568  & 4.935      & 2.339    \\ \hline
16          & 2.719  & 3.827      & 2.940    \\ \hline
17          & 3.169  & 3.958      & 2.613    \\ \hline
18          & 3.307  & 5.083      & 2.392    \\ \hline
19          & 2.422  & 4.985      & 2.480    \\ \hline
{\bf Mean}  & 2.985  & 4.622      & 2.970    \\ \hline
{Std. Dev.} & 4.014  & 13.738     & 18.914   \\ \hline
\end{tabular}
\caption{The average separation between clusters in FormOf, HasContext and SymbolOf, along with their mean and standard deviation. Separation is measured by the distances from the center of a cluster to all other clusters' centers. The cluster IDs in this  Table \ref{table: average_cohesion} are aligned with those in the previous table, for any given relation (e.g., FormOf Cluster 1 in this table is the same as  FormOf Cluster 1 in Table \ref{table: average_cohesion}.}
\label{table: average_separation}
\end{table}

As the name suggests, cohesion measures the extent to which the points in each cluster are tightly grouped together. A cluster with high dispersion would have low cohesion. While this intuitive measure could be quantitatively measured in several conceivable ways, we use a relatively simple formula. First, for each cluster, we calculate the average Euclidean distance between the normalized points in the cluster and the cluster's centroid (also normalized\footnote{Since the graph embeddings are not normalized to lie on a unit-radius hypersphere, we normalize the vectors before computing the distance to enable cross-cluster comparisons, and also comparisons with the (subsequently described) separation measures.}). Since smaller distances indicate greater cohesion, we subtract the average from 1 to obtain a cohesion on a scale of 0.0 to 1.0, with 1.0 indicating perfect cohesion (i.e., all points inside the cluster coincide after normalization). In Table \ref{table: average_cohesion}, we report the cohesion per cluster for each of the relations. Based on the table, we find that the average cohesion for FormOf, HasContext and SymbolOf is 4.439, 4.562 and 4.831 respectively. Note that the absolute values here are not meaningful, and must be interpreted relative to each other. We find that, while the average cohesion scores of clusters in these three relations are close in value, the standard deviations show significant differences. The standard deviations of HasContext cluster cohesion scores are generally higher than the standard deviations of the other two relations' cohesion scores. This simple result suggests the hypothesis that HasContext is more diverse (and hence, more dispersed in embedding space) than the other two relations. It is interesting that the deviation is inversely related to the number of triples corresponding to each relation (but not linearly). 

While cohesion is a good measure for computing the quality of clustering, it is not adequate by itself. An optimal cohesion can be obtained by assigning each point to its own cluster (in which case, the point becomes the centroid of the cluster). An additional metric, even after controlling for $k$, is the separation of the clusters i.e., how `far apart' the different clusters are in the embedding space. Similar to cohesion, there are multiple mathematical ways to capture this qualitative notion. We employ a simple method that is analogous to the cohesion measure--namely, for a given cluster, we compute its separation by computing the average (normed Euclidean) distance from its centroid to each of the other 19 centroids. Table \ref{table: average_separation} reports the results, along with the mean and standard deviation. We find that, once again, HasContext has highest average separation which is 4.622. This suggests that the `contexts' represented by these clusters are well-separated. FormOf and SymbolOf obtain similar average separation (2.985 and 2.970, respectively). In contrast to cohesion, the standard deviation of cluster separation scores is highest for the SymbolOf  relation. 

In comparing the cohesion and the separation of clusters for all three relations in Tables \ref{table: average_cohesion} and \ref{table: average_separation}, we find that the HasContext average separation is close to the average cohesion. In other words, the average distance from one cluster center to a within-cluster node is close to the average distance from that cluster center to other cluster centers. FormOf and SymbolOf's average separation is lower than average cohesion. It suggests that (some) nodes in these two relations' clusters might be closer to other cluster centers than their own centers. Substructures in these two relations are less independent than in HasContext.
\begin{figure}[htbp]
    \captionsetup{font={small}}
    \centering
    \includegraphics[width=9cm]{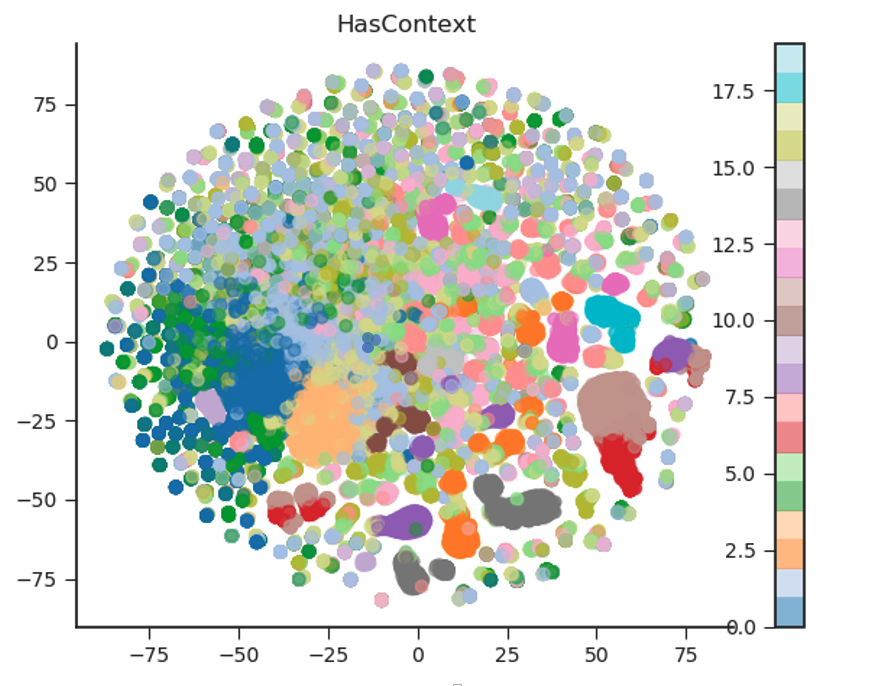}
    \caption{\csentence{The visualization of clusters in HasContext by t-SNE}
      Here, 20 HasContext triples-clusters, discovered using k-Means, are represented in different colors and the ID of each cluster is shown in the color bar on the right side. The cluster IDs are consistent with those introduced and used in Tables \ref{table: HasContext} (in Appendix D), \ref{table: average_cohesion}, and \ref{table: average_separation}. The t-SNE `dimensions' merely work for visualization and do not have intrinsic meaning.
      % According to different colors shown in the plot, its corresponding cluster index could be found in the color bar on the right side and then the example triples in the cluster could be checked in related sample tables.
      }
    \label{figure: hascontext}
  \end{figure}

\begin{figure}[htbp]
    \captionsetup{font={small}}
    \centering
    \includegraphics[width=9cm]{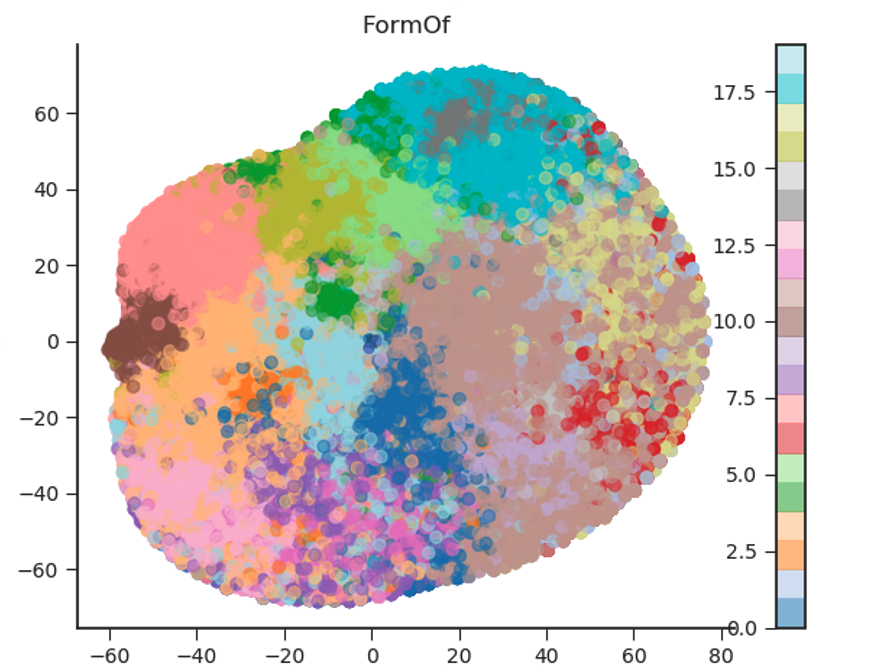}
    \caption{\csentence{The visualization of clusters in FormOf by t-SNE}
      The same methodology applies as in Figure \ref{figure: hascontext}.}
    \label{figure: FormOf}
  \end{figure}

\begin{figure}[htbp]
    \captionsetup{font={small}}
    \centering
    \includegraphics[width=9cm]{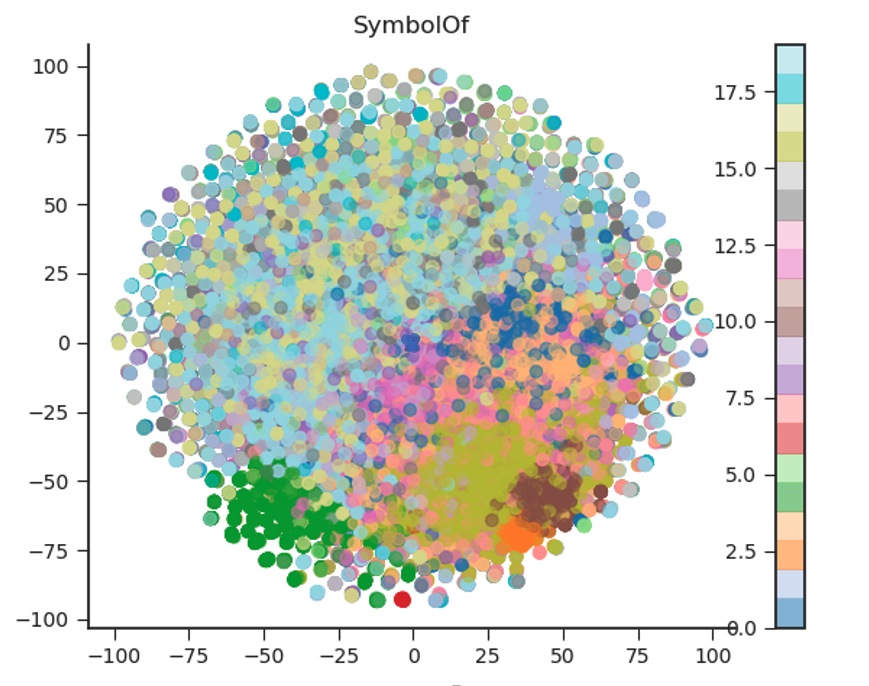}
    \caption{\csentence{The visualization of clusters in SymbolOf by t-SNE}
      The same methodology applies as in Figure \ref{figure: hascontext}.}
    \label{figure: SymbolOf}
  \end{figure}

While these results demonstrate evidence of some substructure for all three relations, it is also instructive to study the substructure by further analyzing triples within these clusters. While a full qualitative study of this nature for all three relations is beyond the scope of this article, we perform a limited study for the HasContext relation, which has particularly `pure' clusters (Figure \ref{figure: hascontext}), by randomly sampling five triples per cluster and determining if we can deduce the `theme' of the cluster from these five triples. We also create the visualization for the substructures of FormOf and SymbolOf in Figures \ref{figure: FormOf} and \ref{figure: SymbolOf}, and sample 5 example triples for each cluster in Tables \ref{table: FormOf} and Figure \ref{figure: SymbolOf_example} in Appendix D, but we do not do an analysis for the sampled triples (for each cluster) of these two relations here.

As illustrated in Table \ref{table: HasContext} in Appendix D (which lists our sampled triples per cluster), the triples show that some of the clusters clearly embody scientific `domains' of study such as geography (Cluster 14), chemistry (Cluster 6), medicine (Cluster 17), mathematics (Cluster 18), and physics (Cluster 19). However, there are also `mixed' clusters that seem to be related to more than one theme (at least on the surface). For example, Cluster 3 contains some `locality' triples, even though Cluster 15 is predominantly concerned with localities, and Cluster 3 is mainly concerned with zoology. We believe that there could be two causes for such `confusion'. The first is due to the automatic and unsupervised nature of the embedding and the second is due to ConceptNet itself, both in terms of the noise within ConceptNet and also because some tail entities (such as \emph{/c/fr/localites}) may be imbalanced in terms of the head entities associated with them. Some other clusters also include some interesting combinations, such as Cluster 12 which contains triples corresponding to both `computing' and `slang'. In the embedding and clustering process, `computing' and `slang'-related triples are thought to be close to each other in embedding space, even though the semantic similarity may not be evident. Yet other clusters (such as Cluster 0) seem to encapsulate the broad notion of HasContext and do not have an evident thematic classification that we can determine.

In using visualization to investigate the inter-relationships between these clusters in Figure \ref{figure: hascontext}, we uncover other interesting findings. For example, the cluster focused on chemistry-related triples seems to overlap with the cluster containing biology (and also astronomy-related triples). While some of the overlap in the figure is exaggerated due to dimensionality reduction, it is nonetheless indicative of the low separation between these two clusters in high-dimensional space and an indirect acknowledgement of the shared lineage of some of these scientific disciplines. An interesting avenue for future exploration could be to quantify and explain such topical overlap between certain clusters by using techniques such as hierarchical clustering.

\subsection{Study 3: Correlation between a specific relation and its negation}
\label{negation}
While Study 1 focused on the `macroscopic' structure of ConceptNet through a study of relational structure, and Study 2 attempted to map specific substructures in three coarse-grained relations, they provided less insight on the relations \emph{between} relations. One such relation is \emph{negation}. Negation is an important everyday relation that also has a distinctive role in formal and logical paradigms (including both propositional and first-order logic). In a mathematical framework, a set $S$ and its negation $\neg S$ are considered to be as different as two sets could get, since they are guaranteed to be disjoint. Another way to think about negation is as a \emph{meta-relation}. While there is a wide body of work on such meta-relations in formal paradigms such as logic [x], not much is known about them with respect to commonsense reasoning. 

In this section, we attempt to derive more structural insights on the nature of the negation relation by studying two specific relations in ConceptNet: \emph{Desires} and \emph{NotDesires}. Since both relations can be explicitly asserted in facts in ConceptNet, we can use these collections of assertions to detect high-level patterns in usage. One question that arises, for example, is: in embedding space, do the \emph{Desires} and \emph{NotDesires} translation embeddings\footnote{Recall, from previous sections, that a translation embedding can be generated per fact or assertion.} neatly separate into clusters? Also, when it is `unknown' whether there is a Desires or NotDesires relation between two entities $A$ and $B$, how does a hypothetical `Unknown' relation compare to Desires or NotDesires? 

% In experiment 3, we want to explore the finding mentioned in section \ref{embedding} that there exists high similarity between a relation and its negation when we visualize the similarity between two relations in the heatmap. Our main research relation is "/r/Desires" and its negation "/r/NotDesires".

To answer these questions, we set up the experiment as follows. Let us define $\mathcal{T}$ as the set of triples containing either the Desires or NotDesires relation. Let $H$ and $T$ be the sets of head and tail entities in $\mathcal{T}$.  
% In other words, $\mathcal{T} \subseteq H \times T =  \{(h,t)| h \in H, t \in T, (h,Desires,t) \in \mathcal{K} \vee (h,NotDesires, t) \in \mathcal{K} \}$, with $\mathcal{K}$ being the knowledge base of ConceptNet triples.
Furthermore, let us define the set $U$ as the subset of $H \times T$ such that neither $(h, Desires, t)$ nor $(h, NotDesires, t)$ is in $\mathcal{T}$. $U$ is precisely the set of `unknown' head-tail entity pairs we intuited about earlier i.e. the combinations of head and tail entities that \emph{individually} participate in at least one Desires/NotDesires relation as a head or tail entity, but that have never been asserted together in a single triple (either with the Desires or the NotDesires relation). 
% Intuitively, $U$ is the set of head/tail entity combinations for which the relation is \emph{unknown} i.e. we do not know whether a head entity in $U$ desires or not desires a tail entity in $U$. 

Given these notions, let us define $H_U$ and $T_U$ as the respective sets of all head and tail entities in $U$. Note that, empirically, we found that $H_U$ is actually equal to $H$ and $T_U$ is equal to $T$ in ConceptNet for the reason that for any given head entity in $\mathcal{T}$, there is at least one tail entity in $\mathcal{T}$ such that we do not know the relationship between $H$ and $T$ (and analogously for the tail entities $T$).

With these notations and definitions in place, we investigate the differences between pairs of entities that are related through Unknown, Desires and NotDesires relations. Each of these three relations yields a set of mutually disjoint\footnote{We found that there were 16 triples where a head-tail entity pair simultaneously participated in both a Desires or NotDesires relation. We remove these contradictory triples, which are small in number to the total number of Desires and NotDesires triples. We also remove duplicate triples, where applicable.} entity pairs. One issue that arises due to the Unknown class is the number of $(h,t)$ pairs in $U$. Since $U = |H \times T| - |\mathcal{T}|$, and since $|H|=2,683, |T|=5,549$, $|\mathcal{T}|=8,352$,  $|U|=14,879,615$. Clearly, visualizing all triples in $U$ is not feasible, nor desired, since $U$ dwarfs $\mathcal{T}$ in size (intuitively, much more is unknown about what the entities desire or not desire, than is known). %{\bf We'll provide a justification for sampling later on based on this.}
To compare all three relations on a relatively balanced footing, we sample relation triples from $U$ as follows. To obtain roughly the same number of samples as in $\mathcal{T}$, we set a \emph{tail sampling ratio}, $r=\lceil |\mathcal{T}|/2 * |H_U|\rceil$ rounded down to $r'= 2$ i.e. for each head entity $h$ in $H_U$, we randomly sample $r'$ tail entities from the set $\{t|(h,t) \in U \}$, if $r'$ is less than the cardinality of this set (otherwise, we sample the whole set). After sampling, we get a sampled set $U' \subset U$, with $|U'|=5,366$.

Next, we embed $\mathcal{T} \cup U'$ following the translational methodology that we described earlier for the previous research questions, and assign one of three colors to each embedding, based on whether the embedding represents a pair in $U'$, or (if the pair is in $\mathcal{T}$), whether it represents a Desires/NotDesires triple.
% \footnote{We also checked for inconsistencies, namely whether an entity pair $(h,t)$ both participates in a Desires and NotDesires relation. Actually, there are 16 triples simultaneously appear in these two relations owing to the complexity of semantic environment, for example `/c/en/fall /r/Desires  /c/en/season' and `/c/en/fall /r/NotDesires /c/en/season'. Fall has two main definitions as noun, one is the season when leaves fall from trees which desire entity "season" and the other is a falling out, off, or away, like waterfall, which not belongs to a season. These edges have been removed from visualization. Along with the removal of two relation triple's intersection set, duplicates of some triples have also been filtered. Before filtration, Desires and NotDesires have 8625 triples in total.}. 

\begin{figure}
\captionsetup{font={small}}
\centering

%\begin{minipage}[t]{0.45\textwidth}
\centering
\includegraphics[width=5.8cm]{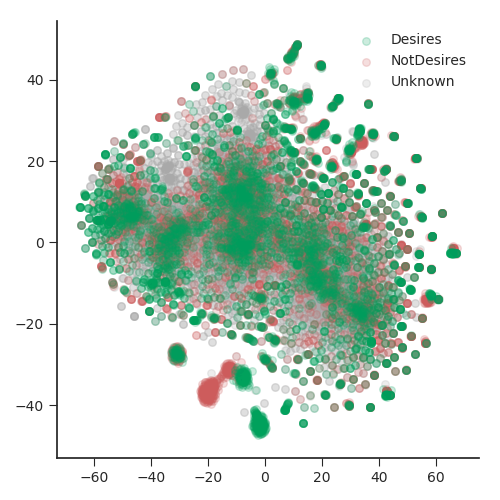}
\caption{\csentence{The visualization of Desires, NotDesires, and Unknown triples using t-SNE}
      The methodology for constructing Unknown triples is described in the text. Data points corresponding to Desires, NotDesires and Unknown triples are shown in green, red and gray, respectively. Similar to previous t-SNE visualizations, the dimensions are only for visualization and do not have intrinsic meaning. Since the points nearly coincide in this 2D space, we `overlay' the Desires points on the NotDesires points where there is a conflict (with the opposite result shown in the next figure)}
\label{figure: notdesires&desires}
\end{figure}

\begin{figure}
\captionsetup{font={small}}
\centering
%\end{minipage}
%\begin{minipage}[t]{0.45\textwidth}
\centering
\includegraphics[width=5.8cm]{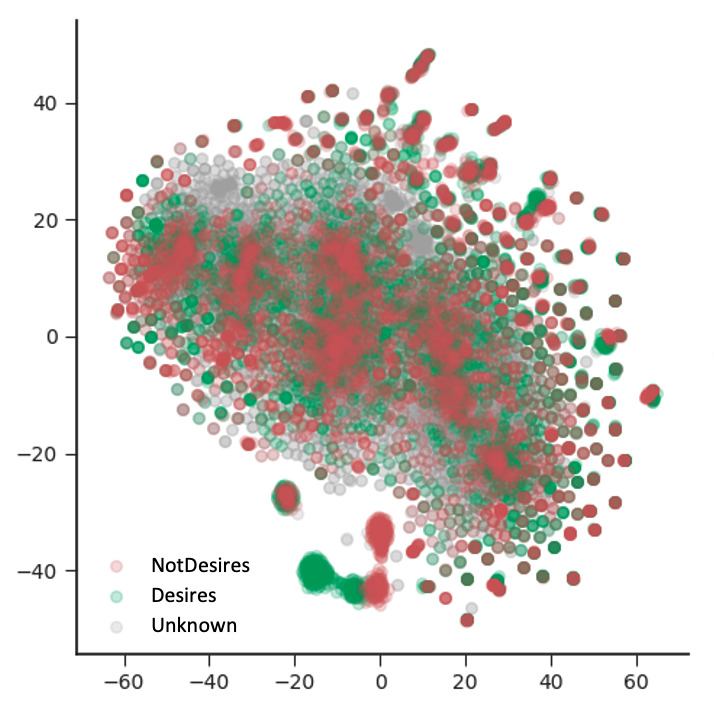}
\caption{\csentence{The visualization of NotDesires, Desires, and Unknown triples using t-SNE}
      The methodology is the same as in Figure \ref{figure: notdesires&desires}, but in this figure, we `overlay' the NotDesires points when there is a conflict.}
\label{figure: desires&notdesires}

%\end{minipage}
\end{figure}

Contrary to expectation, the triples in Desires and NotDesires do not separate in the t-SNE visualization. To  better observe the relationship, we render the same visualization in two different ways (Figures \ref{figure: notdesires&desires} and \ref{figure: desires&notdesires}). In one of these figures, the Desires translational embeddings are `overlaid', while in the other, the NotDesires embeddings are overlaid. Regardless of how we visualize the triples, we find that there is high overlap between them. The primary reason clearly is that the same \emph{semantic categories} of head and tail entities tend to participate in both Desires and NotDesires relations. For example, a person may Desire a movie, but may NotDesire another movie, which would lead to the kind of conflation that we observe in Figures \ref{figure: notdesires&desires} and \ref{figure: desires&notdesires}.  Hence, the nuances between these two relations are lost in such a plot. 

The Unknown embeddings also yield interesting conclusions. While these embeddings also have high overlap with the other embeddings (the `middle' portion of Figures \ref{figure: notdesires&desires} and \ref{figure: desires&notdesires}), they also occupy portions of the embedding space not occupied by either \emph{Desires} or \emph{NotDesires}. For example, there is a long grey lobe both at the top and the bottom of Figures \ref{figure: notdesires&desires} and \ref{figure: desires&notdesires}. These are precisely the types of spaces that \emph{link prediction} (and other recommendation-style) algorithms attempt to map using a set of semantic and structural features \cite{link-prediction1}, \cite{link-prediction2}. 

Nevertheless, there are alternative ways to detect such nuances, if they exist. As proof-of-concept, we trained ordinary machine learning classifiers (both random forest and logistic regression) to determine whether the embeddings still contain latent information (that may not easily be visualized) about the Desires and NotDesires relations. Specifically, we used 10-fold cross-validation to train and evaluate a classifier on predicting whether a translational embedding could be correctly labeled as Desires or NotDesires. The best performance was achieved with a tuned random forest model that obtained an accuracy of 76\% on this task. Hence, the negation meta-relation does leave behind a semantic trace that a machine learning model, appropriately trained, is able to learn, but that is not discernible in an unsupervised framework such as clustering or visualization\footnote{The t-SNE visualization algorithm may itself be thought of as a kind of clustering based on optimizing relative distances in a `visual' i.e. 2-D space.}. Hence, there are semantic differences between the two relations, but it is an open question whether there is a \emph{structural} difference, structure being defined similarly as in Study 2. 

To conclude, this study clearly refutes the hypothesis that a relation like Desires should have a `clear' separation from its negated version. Note the difference from Study 2, where we found clear separations and substructures upon analyzing triples that belonged to a \emph{single} relation that had one definition (at least in principle). The qualitative and structural differences between Unknown and the other two relations provides evidence of the somewhat intuitive notion that unknown relations constitute their own relation class in such commonsense knowledge bases. This is an important aspect of ConceptNet that commonsense reasoners need to account for when using it as a distant supervision resource for question answering tasks. In fact, a key problem that commonsense reasoners need to successfully address is accurate inference over `unknown'-relation spaces (not only for Desires, but other relations of a similar nature) when presented with questions or scenarios that require such inference.

\section{Discussion}
\label{discussion}
The results in Sections \ref{embedding}, \ref{substructure}, and \ref{negation} reveal some insights into commonsense knowledge through an empirical and systematic study of ConceptNet's relational structure (and substructures of certain relations), as well as the characteristics of the ubiquitous \emph{negation} relation. Except the experiment results which have been detailed presented above, there are still some findings needed to be mentioned. We summarize these findings below:

\begin{enumerate}
    \item By using a methodology based on centroid relation vectors (Section \ref{embedding}), we were able to magnify the (sometimes, subtle) differences between relations  using a symmetric heatmap. In contrast, using `direct' relation vectors was found to lead to more uniformity. Similarly, in considering other methodologies for detecting relational similarity, we found that comparing explicit textual definitions had limited use, and when considering overlap between entities (rather than graph embeddings), overlap between head entities-sets of two relations had more utility and intuitive plausibility than overlap between the relations' tail entities-sets. 
    % \item In exploring the substructures of three specific relations (SymbolOf, FormOf, and HasContext), we found that... 
    \item We also proposed and used a methodology for studying a relation (Desires) in the context of its negation (NotDesires), and a novel `Unknown' relation that expressed (given a head entity and a tail entity) that we did not know whether the relation or its negation applied to the two entities. We found that, while the Unknown relation seems to occupy a distinct portion of the embedding space, the Desires and NotDesires embeddings overlap significantly. In terms of usage in assertions (in the ConceptNet KG), the Desires and NotDesires relations are not very different. Yet, the two are also not indistinguishable, since machine learning methods like random forest are able to distinguish between them (when trained on some labeled assertions). We believe that a similar finding could apply to other such relation `pairs', such as causal relations and sub-relations (e.g., HasSubevent and HasLastSubevent). 
\end{enumerate}

We have also empirically observed that, while `super-class' semantics tend to be associated with the definition of HasContext (i.e., when head entity A is a word that is used in the context of tail entity B, B tends to be a more general, abstract `super-class' of A, such as a topic area, technical field, or regional dialect, as is also mentioned in the official definition of HasContext), there are significant substructures that can't be uniformly explained by an umbrella term like HasContext. These substructures may help us better understand what the different contexts are in which people interpret pairs of words or entities. Understanding context is critical for building systems that have commonsense, such as chatbots and conversational agents that are able to understand sentences in the specific context in  which the sentences are uttered.

A particularly interesting relation is  SymbolOf. The ratio of the size of the head entity set of SymbolOf to the size of tail entity set is 0.011, almost 1/200 of the average value observed for all relations. This indicates that the number of words or phrases which could be used to `explain' the symbols is far greater than the number of symbols themselves. Symbols (and arguably, by extension, the emojis used on social media) are necessarily under-determined, and the semantics assigned to symbols vary in different contexts. While not qualitatively unsurprising, our results suggest that the `diffusion' of symbol-semantics is far greater than one might have thought. 

% Although it is common that one symbol could have multiple textual expressions and one word or phrase could also be symbolized in multiple ways,  we don't have a clear sense of the ratio between them. We are even uncertain about which one has a larger size, the set of words which could be symbolized or the set of symbols of words. However, it is clear that, according to the knowledge in the current version of ConceptNet, the number of symbols is much less than the number of words or phrases which could be used to explain the symbols. A plausible explanation is that each symbol tends to have developed various meanings based on its original meaning.

There are other interesting findings about the ratio of the number of unique head entities (participating in a  relation) to the number of tail entities, although these might be more specific to ConceptNet than to commonsense knowledge per se. For example, the ratio for both  HasFirstSubevent and HasLastSubevent relations is around 0.3, which indicates that, on average (in ConceptNet), a `subevent' has the potential to be the first in three ($\approx 1/0.3$) sequences of subevents (that each constitute an `event'). A similar conclusion applies for the terminal subevent in the sequence. Whether this is a general statistical truth or not is left for future work to explore.
% and the number of ending options is a little higher than the initialization options. We leave these findings for future research.

\section{Conclusion}

In this article, we explored a set of data-driven methodologies, using the ConceptNet knowledge graph, to derive insights at scale on the nature of commonsense knowledge. To this end, we conducted a set of three carefully designed studies, and a range of experiments. Our results show that there is significant structure (and substructure) in ConceptNet (Study 1), especially with respect to relations such as FormOf, SymbolOf and HasContext. Study 2, in particular, shows that there are at least (and possibly more than) 20 different kinds of context that can be discovered within ConceptNet, some of them very well-defined (such as a scientific field of study, or even slang), with others being more diffuse. In some cases, there are unusual but non-random degrees of overlap and association (e.g., computer science and slang). In Study 3, we apply a different, but related, methodology, to study negation, and find that `unknown'-type relations occupy unique regions of the embedding space, whereas the two relations (Desires and NotDesires) significantly overlap in the embedding space, even though a trained machine learning model can distinguish between them on the basis of their features. 

Several avenues of future research remain. For example, it may be worthwhile going even deeper into a relation like HasContext to discover if there are \emph{hierarchical} substructures, rather than a single layer of substructures. Hierarchical clustering algorithms in the graph embedding space could be used to achieve this goal methodologically. Similar to Study 3, a relation and its causal equivalent, could be studied together to understand their connection in an embedding space. A more ambitious line of study would be to try and connect these empirical results to the more theoretical results on commonsense reasoning (e.g., by \cite{csbook1}).

\begin{backmatter}

\section*{Competing interests}
  The authors declare that they have no competing interests.

\section*{Author's contributions}
Shen performed the experimental study and contributed significantly to writing and study design. Kejriwal primarily contributed to the writing and project ideation, as well as formulation of the specific research questions. 
%     This paper proposes an in-depth study corresponding to the detailed  relational  structure  in  embedding  space, including distributions and correlations of ConceptNet entities and relations in low-dimension embedding space and validates the efficiency of information packed in these embeddings by machine language models and graph analysis. The experiment result shows several meaningful and interesting patterns. Our contributions could be summarized as follow: 
% \begin{itemize}
% \item[*] We study the relational structure of ConceptNet by using a recently released, state-of-the-art graph embedding framework to embed the nodes and relations in ConceptNet  into  a  dense,  continuous  vector  space,  and  present  well-defined techniques for classifying and studying relational structure in the context of this space.
% \item[*] For specific high-volume and coarse-grained relations (specifically,  HasContext,  SymbolOf  and FormOf), we explore substructures by clustering model and graphic analysis, and obtain fine-grained sub-relations. 
% \item[*] We contrast structural relation between an important relation 'Desires' and its negation 'NotDesires', which finds large-scale overlaps and triples differences in overlap could be easily learned by machine learning model.
% \end{itemize}

\section*{Acknowledgements}
  This project was funded under MOWGLI, a project in the Defense Advanced Research Projects Agency (DARPA) Machine Common Sense program, supported by United States Office Of Naval Research under Contract No. N660011924033. The views expressed in this paper are those of the authors. \ldots
%%  After submission of the .TEX file,                     %%
%%  you will be prompted to submit your .BBL file.         %%
\section*{Appendix A: Validating PBG Embeddings Using KL-Divergence}
Given two probability distributions $p$ and $q$ with the same support (intuitively, the same event space or `x-axis' over which the probability density/mass function is computed), the KL-Divergence $D_{KL}(p||q)$ of $p$ with respect to $q$ is given by:
\begin{equation}
D_{KL}(p||q) = \sum^N_{i=1}p(x_i) \cdot (\log p(x_i) - \log q(x_i))
\end{equation}

\begin{table}[h!]
\begin{tabular}{|c|c|c|c|}
\hline
{\bf Relation}                  & {\bf KL-divergence} & {\bf Relation}                & {\bf KL-divergence}         \\ \hline
IsA                       & 0.759         & NotDesires              & 1.512                 \\ \hline
dbpedia/knownFor          & 0.455         & PartOf                  & 1.015                 \\ \hline
HasSubevent               & 0.962         & dbpedia/genus           & 0.589                 \\ \hline
Entails                   & 0.782         & EtymologicallyRelatedTo & 0.630                 \\ \hline
DerivedFrom               & 1.109         & HasA                    & 0.472                 \\ \hline
UsedFor                   & 1.470         & Desires                 & 1.085                 \\ \hline
CapableOf                 & 0.924         & dbpedia/leader          & 0.286                 \\ \hline
AtLocation                & 0.209         & CreatedBy               & 0.143                 \\ \hline
HasContext                & 0.935         & NotUsedFor              & 0.682                 \\ \hline
Antonym                   & 0.662         & DefinedAs               & 0.589                 \\ \hline
HasLastSubevent           & 1.367         & SymbolOf                & 0.711                 \\ \hline
CausesDesire              & 1.252         & LocatedNear             & 0.308                 \\ \hline
EtymologicallyDerivedFrom & 1.103         & HasPrerequisite         & 1.141                 \\ \hline
InstanceOf                & 1.290         & MadeOf                  & 0.386                 \\ \hline
dbpedia/influencedBy      & 0.751         & ReceivesAction          & 1.033                 \\ \hline
MannerOf                  & 1.669         & dbpedia/capital         & 0.493                 \\ \hline
dbpedia/language          & 0.854         & Causes                  & 0.725                 \\ \hline
HasProperty               & 0.129         & NotHasProperty          & 0.477                 \\ \hline
dbpedia/product           & 0.657         & NotCapableOf            & 0.330                 \\ \hline
HasFirstSubevent          & 0.940         & dbpedia/field           & 1.209                 \\ \hline
dbpedia/genre             & 0.720         & SimilarTo               & 0.264                 \\ \hline
DistinctFrom              & 1.056         & MotivatedByGoal         & 0.121                 \\ \hline
dbpedia/occupation        & 1.603         & ObstructedBy            & 0.797                 \\ \hline
FormOf                    & 1.401         & RelatedTo               & 0.999                 \\ \hline
Synonym                   & 1.365         & \multicolumn{1}{l|}{}   & \multicolumn{1}{l|}{} \\ \hline
\end{tabular}
\caption{The KL-Divergence of 49 relations when using the per-relation similarity lists and methodology described in the text. The two probability distributions used in the KL-divergence calculation for relation $r$ (Equation 2) are the frequencies of cosine similarity scores in the two lists ($SL_r$ and $SL_r^{'}$).} 
\label{table: KL-divergence}
\end{table}

The higher the KL-Divergence, the more different the two distributions are. Note also that the measure is asymmetric in the general case i.e. $D_{KL}(p||q) \neq D_{KL}(q||p)$.

In our particular setting, we want to use KL-Divergence to measure how much information is `lost' when we use $\vec{r_c}$ as a replacement for $\vec{r}$. If not much information is lost (i.e. the KL-Divergence is low), then $\vec{r_c}$ may be said to be a \emph{robust} replacement for $\vec{r}$. Specifically, given the lists $SL_r$ and $SL_r^{'}$ for relation $r$, we first compute an empirical probability distribution for each list by using frequencies of scores in the lists. Next, we compute the KL-Divergence by using the probability distribution of $SL_r$ as the reference distribution $q$. For each relation, the KL-Divergence is tabulated in Table \ref{table: KL-divergence} and we show two examples in Figure \ref{figure:KL-divergence}. Both the table and figure show that divergence (between using the centroid and the `directly output' vector) is generally low, attesting to the previous result the PBG embeddings over the 4 million triples-sample, and our methodology for constructing the centroid embedding per relation, are consistent and robust.
\begin{figure}
    \captionsetup{font={small}}
    \centering
    \begin{minipage}[t]{0.45\textwidth}
    \centering
    \includegraphics[width=6cm]{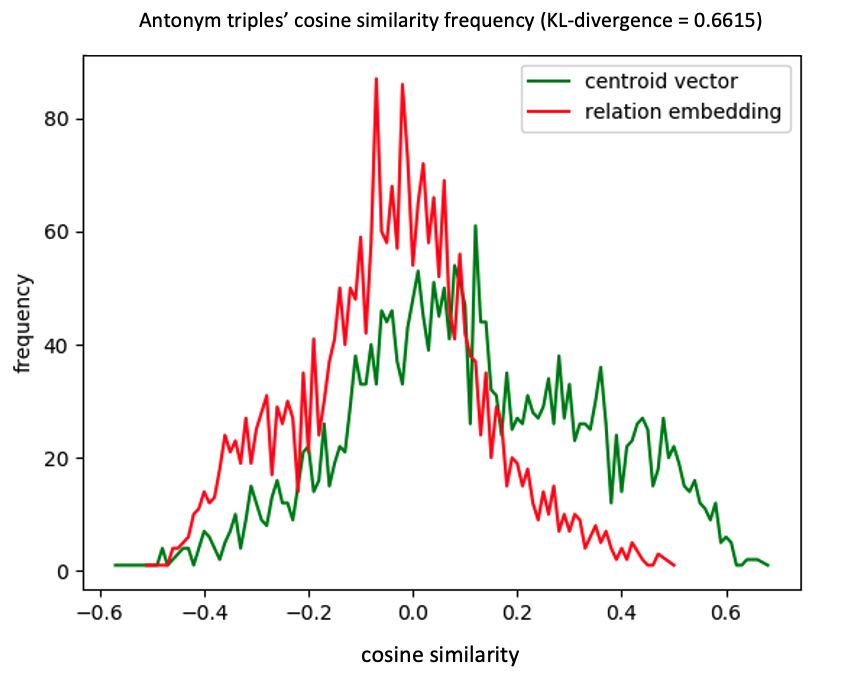}
    % \caption{negative example(Antonym)}
    \end{minipage}
    \begin{minipage}[t]{0.45\textwidth}
    \centering
    \includegraphics[width=6cm]{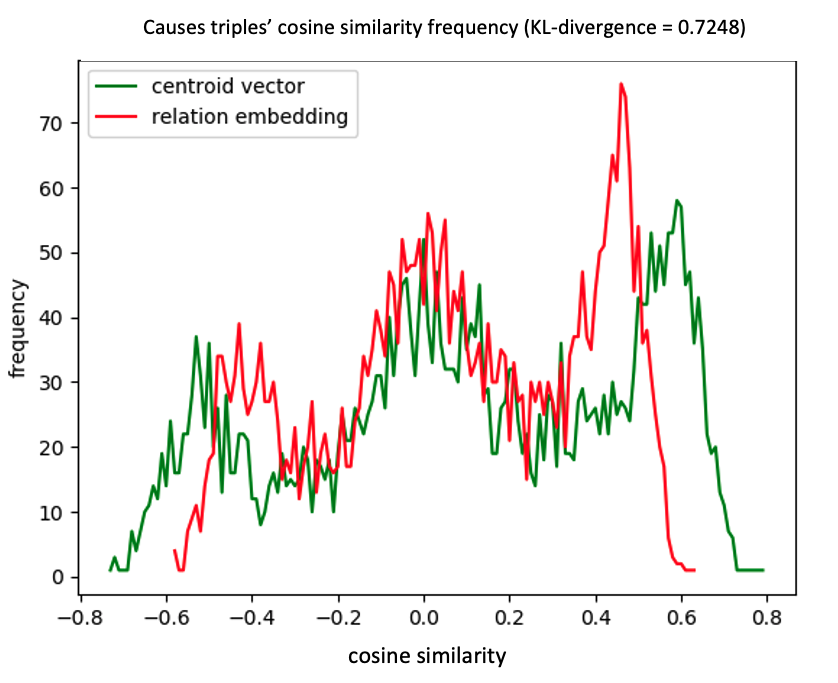}
    % \caption{positive example(Causes)}
    \end{minipage}
    \caption{\csentence{The cosine similarity frequency line chart} 
    The frequency distribution of cosine similarities for two relations \emph{Antonym} (left) and \emph{Causes} (right), where the similarity is calculated between the translational embedding of a triple (containing the relation) and one of two vectors used to represent the relation (centroid vector, or direct relation embedding output by PBG). For each of the two relations, and for each choice of the relation representation, a `cosine similarity' distribution is thereby obtained with each triple containing the relation contributing a single data point to the distribution.}
    %A line plot describes the similarity frequency dispersion between each triple's translational embedding with its centroid vector and real relation embedding for relation Antonym(left) and Causes(right), the KL-divergence score is shown in figure title.}
    \label{figure:KL-divergence}
  \end{figure}
  
  \section*{Appendix B: Definitions of the Main Relations in ConceptNet5}
  
  We reproduce the definitions of the 33 `main' relations (except ExternalURL) in ConceptNet5 from the original project page\footnote{\url{https://github.com/commonsense/conceptnet5/wiki/Relations}}, for the sake of completeness:
  
  \begin{enumerate}
  \item {\textbf{RelatedTo}: The most general relation. There is some positive relationship between A and B, but ConceptNet can't determine what that relationship is based on the data. This was called "ConceptuallyRelatedTo" in ConceptNet 2 through 4. Symmetric.} 
  \item {\textbf{FormOf}: A is an inflected form of B; B is the root word of A.}
  \item {\textbf{IsA}: A is a subtype or a specific instance of B; every A is a B. This can include specific instances; the distinction between subtypes and instances is often blurry in language. This is the hyponym relation in WordNet.}
  \item {\textbf{PartOf}: A is a part of B. This is the part meronym relation in WordNet.}
  \item {\textbf{HasA}: B belongs to A, either as an inherent part or due to a social construct of possession. HasA is often the reverse of PartOf.}
  \item {\textbf{UsedFor}: A is used for B; the purpose of A is B.}
  \item {\textbf{CapableOf}: Something that A can typically do is B.}
  \item {\textbf{AtLocation}: A is a typical location for B, or A is the inherent location of B. Some instances of this would be considered meronyms in WordNet.}
  \item {\textbf{Causes}: A and B are events, and it is typical for A to cause B.}
  \item {\textbf{HasSubevent}: A and B are events, and B happens as a subevent of A.}
  \item {\textbf{HasFirstSubevent}: A is an event that begins with subevent B.}
  \item {\textbf{HasLastSubevent}: A is an event that concludes with subevent B.}
  \item {\textbf{HasPrerequisite}: In order for A to happen, B needs to happen; B is a dependency of A.}
  \item {\textbf{HasProperty}: A has B as a property; A can be described as B.}
  \item {\textbf{MotivatedByGoal}: Someone does A because they want result B; A is a step toward accomplishing the goal B.}
  \item {\textbf{ObstructedBy}: A is a goal that can be prevented by B; B is an obstacle in the way of A.}
  \item {\textbf{Desires}: A is a conscious entity that typically wants B. Many assertions of this type use the appropriate language's word for "person" as A.}
  \item {\textbf{CreatedBy}: B is a process or agent that creates A.}
  \item {\textbf{Synonym}: A and B have very similar meanings. They may be translations of each other in different languages. This is the synonym relation in WordNet as well. Symmetric.}
  \item {\textbf{Antonym}: A and B are opposites in some relevant way, such as being opposite ends of a scale, or fundamentally similar things with a key difference between them. Counterintuitively, two concepts must be quite similar before people consider them antonyms. This is the antonym relation in WordNet as well. Symmetric.}
  \item {\textbf{DistinctFrom}: A and B are distinct member of a set; something that is A is not B. Symmetric.}
  \item {\textbf{DerivedFrom}: A is a word or phrase that appears within B and contributes to B's meaning.}
  \item {\textbf{SymbolOf}: A symbolically represents B.}
  \item {\textbf{DefinedAs}: A and B overlap considerably in meaning, and B is a more explanatory version of A.}
  \item {\textbf{MannerOf}: A is a specific way to do B. Similar to "IsA", but for verbs.}
  \item {\textbf{LocatedNear}: A and B are typically found near each other. Symmetric.}
  \item {\textbf{HasContext}: A is a word used in the context of B, which could be a topic area, technical field, or regional dialect.}
  \item {\textbf{SimilarTo}: A is similar to B. Symmetric.}
  \item {\textbf{EtymologicallyRelatedTo}: A and B have a common origin. Symmetric.}
  \item {\textbf{EtymologicallyDerivedFrom}: A is derived from B.}
  \item {\textbf{CausesDesire}: A makes someone want B.}
  \item {\textbf{MadeOf}: A is made of B.}
  \item {\textbf{ReceivesAction}: B can be done to A.}
  \end{enumerate}

\section*{Appendix C: Additional Analyses and Data in Support of Study 1}

\begin{table}[!h]
    \centering
    \begin{tabular}{|c|c|c|}
        \hline
        {\bf relation}                      & {\bf closest relation}             & score \\ \hline
RelatedTo                  & Synonym                   & 0.057 \\ \hline
FormOf                     & RelatedTo                 & 0.089 \\ \hline
lsA                        & DerivedFrom               & 0.028 \\ \hline
PartOf                     & AtLocation                & 0.052 \\ \hline
HasA                       & AtLocation                & 0.046 \\ \hline
UsedFor                    & HasSubevent               & 0.038 \\ \hline
CapableOf                  & CausesDesire              & 0.038 \\ \hline
AtLocation                 & PartOf                    & 0.052 \\ \hline
Causes                     & MotivatedByGoal           & 0.061 \\ \hline
HasSubevent                & Motivated ByGoa 1         & 0.064 \\ \hline
HasFirstSubevent           & HasSubevent               & 0.041 \\ \hline
HasLastSubevent            & HasFirstSubevent          & 0.01  \\ \hline
HasPrerequisite            & HasSubevent               & 0.017 \\ \hline
HasProperty                & MotivatedByGoal           & 0.017 \\ \hline
MotivatedByGoal            & HasSubevent               & 0.064 \\ \hline
ObstructedBy               & Not Used For              & 0.004 \\ \hline
Desires                    & MotivatedByGoal           & 0.057 \\ \hline
CreatedBy                  & dbpedia/occupation        & 0.012 \\ \hline
Synonym                    & RelatedTo                 & 0.092 \\ \hline
Antonym                    & RelatedTo                 & 0.092 \\ \hline
DistinctFrom               & lsA                       & 0.015 \\ \hline
DerivedFrom                & EtymologicallyRelatedTo   & 0.014 \\ \hline
SymbolOf                   & NotUsedFor                & 0.021 \\ \hline
DefinedAs                  & Antonym                   & 0.001 \\ \hline
ReceivesAction             & Antonym                   & 0.001 \\ \hline
MannerOf                   & Entails                   & 0.032 \\ \hline
LocatedNear                & lnstanceOf                & 0.004 \\ \hline
HasContext                 & PartOf                    & 0.016 \\ \hline
SimilarTo                  & Synonym                   & 0.004 \\ \hline
EtymologicallyRelatedTo    & EtymologicallyDerivedFrom & 0.073 \\ \hline
EtymologicallyDerivedFrom & EtymologicallyRelatedTo   & 0.073 \\ \hline
CausesDesire               & Desires                   & 0.049 \\ \hline
MadeOf                     & Desires                   & 0.033 \\ \hline
    \end{tabular}
    \caption{The most similar relation for each relation by selecting the one that obtains the highest Jaccard similarity when comparing the tail entity sets of two relations.}
\label{table: most_similar_relations_tail}
\end{table}

%\subsection*{t-SNE Visualizations and Clustering Analysis of Relation Embeddings}

In Study 1, we used the KL-Divergence to define `heatmaps' between relation embeddings (using both centroid and direct relation embeddings); however, a more informal analysis could also be undertaken using the t-Distributed Stochastic Neighbor Embedding (t-SNE) algorithm \cite{t-sne}, which is an established neural dimensionality-reduction algorithm that has been shown to be superior to other dimensionality-reduction techniques (e.g., Principle Components Analysis) for visualizing the outputs of representation learning algorithms such as PBG. 

Our methodology is similar to that in Study 1 for generating KL-Divergence heatmaps. Once the centroid vectors are obtained, one for each relation, we visualize the similarities between the relations in two dimensions by using t-SNE (Figure \ref{figure: centroid_distribution}).     
\begin{figure}
  \captionsetup{font={small}}
  \centering
  \includegraphics[width=12cm]{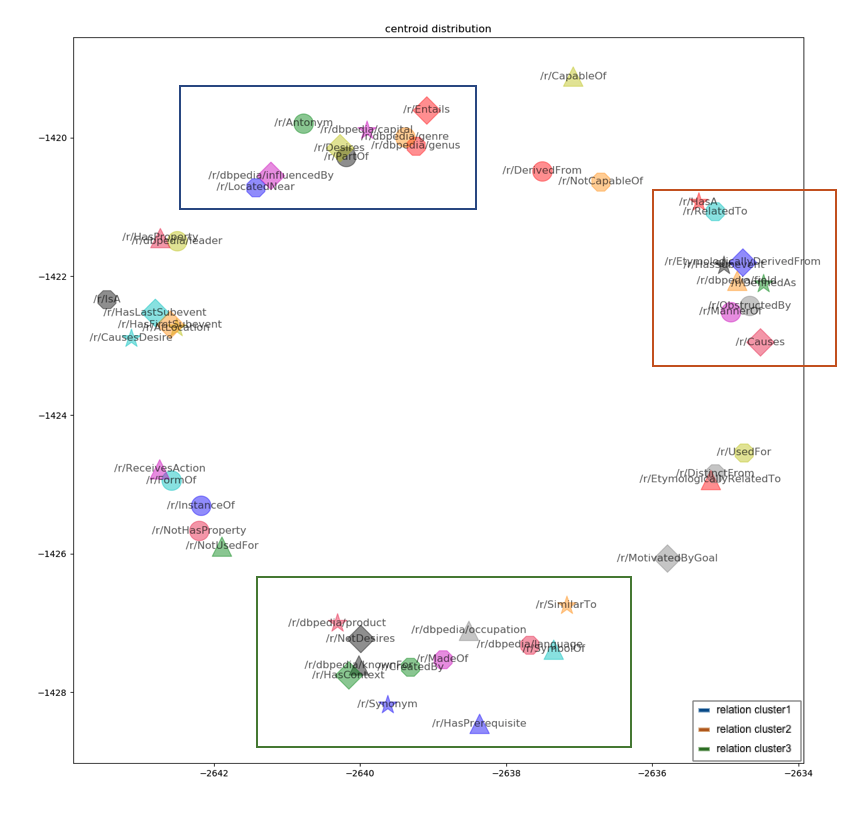}
  \caption{\csentence{The visualization of 49 relation centroid vectors using t-SNE.}
      The centroid vector for each relation is calculated using Equation \ref{eqn:centroid_vector}. Note that t-SNE dimensions (or axes) are only used for visualization and do not have intrinsic meaning. We manually mark the three obvious clusters by the rectangle for further analysis.}
  \label{figure: centroid_distribution}
\end{figure}

The figure shows that, while many relations are reasonably uniformly distributed (visualized in the figure as a `circle' with relatively uniform spacing between relations), some relations do congregate in clusters, suggesting significantly higher inter-similarities in the embedding space. These findings are largely consistent with those reported in the main text. Three of the most prominent clusters are manually demarcated in the figure, and are described below:

\begin{enumerate}
    \item The cluster in the upper left contains the relations Entails, Antonym, Desires, ParOf, LocatedNear, dbpedia/capital, dbpedia/genre, dbpedia/genus, and dbpedia/influencedBy. Although the dictionary semantics of these relations are quite different, the similarities in embedding space shows that (relatively speaking) there is likely high overlap between the sets of entity pairs that are related using each of these relations. For example, LocatedNear is described as `A and B are typically found near each other'; dbpedia/influencedBy states that `A is influenced by B'. If two objects were close to each other, it is quite possible that one object is influenced by another. These kinds of complex interplays seem to be captured, at least in part, in the embedding space and visualization.
    
    \item The cluster in the upper right contains the relations HasA, RelatedTo, EtymologicallyDerivedFrom, HasSubevent, dbpedia/field, DefinedAs, ObstructedBy, MannerOf, and Causes. The large degree of overlap between the corresponding entity sets (the union set of head and tail entity set) is the main reason why relations are grouped in this cluster. The closeness between ObstructedBy and MannerOf is an obvious instance. ObstructedBy indicates that ``B is an obstacle to achieve the goal action A'' and MannerOf emphasizes that ``A is a specific way to do goal action B". The overlap of ``action''-type entities directly results in the proximity of ObstructedBy to MannerOf.
    
    \item The cluster in the bottom contains the relations NotDesires, Synonym, SymbolOf, MadeOf, HasPrerequisite, HasContext, SimilarTo, CreatedBy, dbpedia/product, dbpedia/occupation, dbpedia/language, and dbpedia/KnownFor. This cluster is similar to the first cluster, but contains some negations of relations in the first cluster. One interpretation of this cluster is that it may be the converse or negation of the first cluster, even though the actual relations within this cluster do not seem to be closely semantically related to one another.
\end{enumerate}

An interesting question that arises is whether similar conclusions can be drawn from the relation embeddings that were directly output by the PBG algorithm.  These `direct' relation embeddings are visualized in Figure \ref{figure: relation_distribution}, using the same methodology as for the centroid embeddings. While there are a few similarities between Figures \ref{figure: centroid_distribution} and \ref{figure: relation_distribution}, there are also significant differences. For example, the distribution is more uniform, suggesting that direct relation embeddings may not (hypothetically) capture `usage' semantics as effectively as the centroid embeddings. Again, these findings are consistent with the heatmaps illustrated in the main text for Study 1. In the clusters that are observed, we find that the relations have a more obvious semantic connection to each other (e.g., EtymologicallyDerivedFrom and DerivedFrom, Desires and CausesDesire). Other less obvious examples also point to a similar conclusion.

\begin{figure}[h!]
  \captionsetup{font={small}}
  \centering
  \includegraphics[width=12cm]{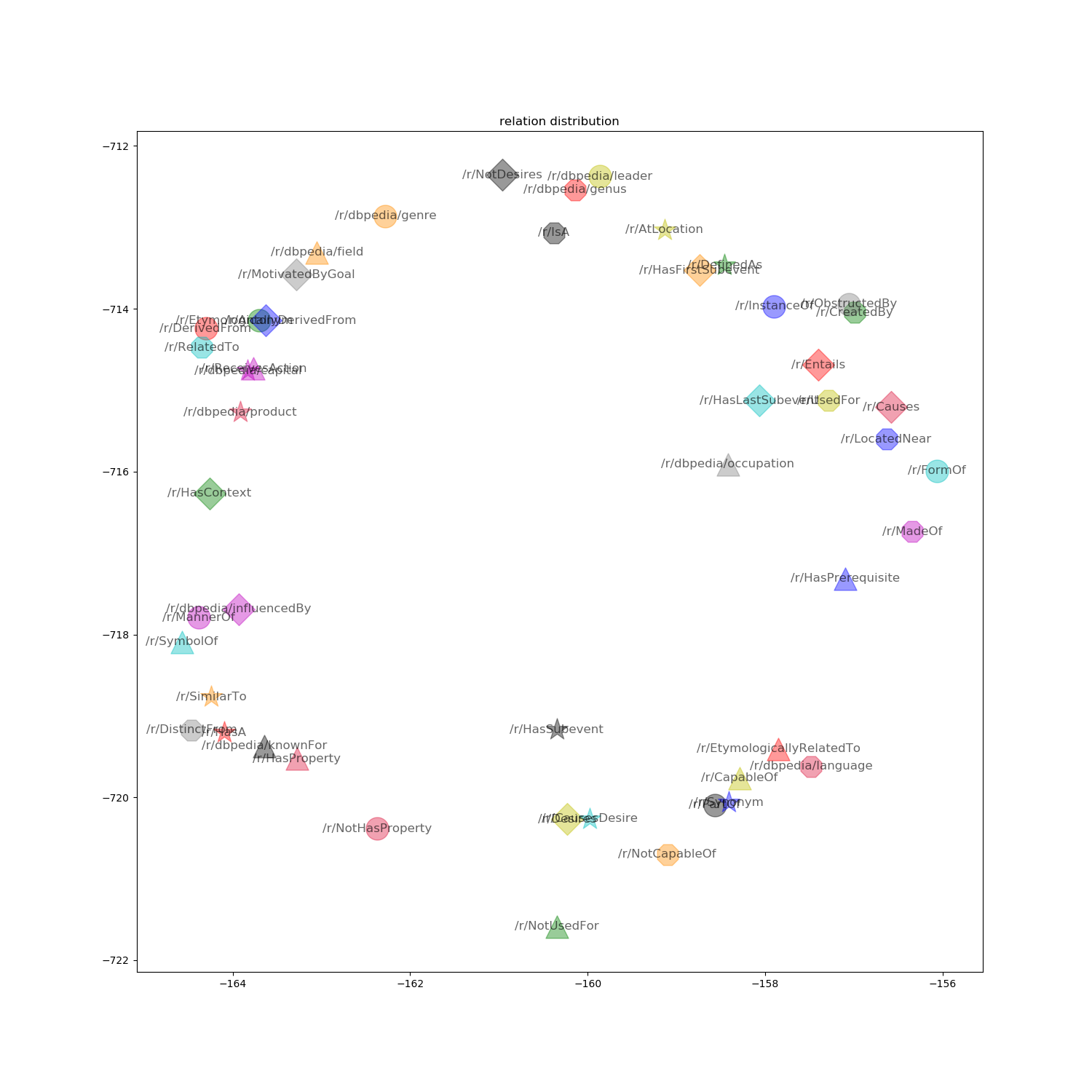}
  \caption{\csentence{The visualization of 49 relation embeddings using t-SNE}
      The relation embeddings are directly yielded by PBG. The methodology is the same as \ref{figure: centroid_distribution}; note that the t-SNE dimensions do not have intrinsic meaning and are only useful for 2D visualization.}
      % The distribution of relation embeddings produced by RobtertA.}
  \label{figure: relation_distribution}
\end{figure}

Despite their differences, the two sets of results (`centroid' and `direct') illustrate that all relations are not equal (i.e. uniformly distributed), and that some structure does exist in the relation embedding space. In the case of centroid embeddings, usage semantics seem to be captured more effectively in the visualization, while in the case of direct embeddings, a more direct semantic connection (learned by the PBG  in the course of training) seems to be necessary for relations to be clustered together. However, there is clear information loss in projecting the embeddings down to 2D. A proper analysis should therefore rely on the information-theoretic methodology in the main text, and not on informal visual analysis, though the visualizations are undoubtedly useful for gaining a `sense' of how the relations differ from each other depending on the embedding methodology employed. 

\section*{Appendix D: Additional Analyses in Support of Study 2}
\begin{figure}[htbp]
  \begin{minipage}[t]{0.31\textwidth}
    \centering
    \includegraphics[width=1.7in, height=2.0in]{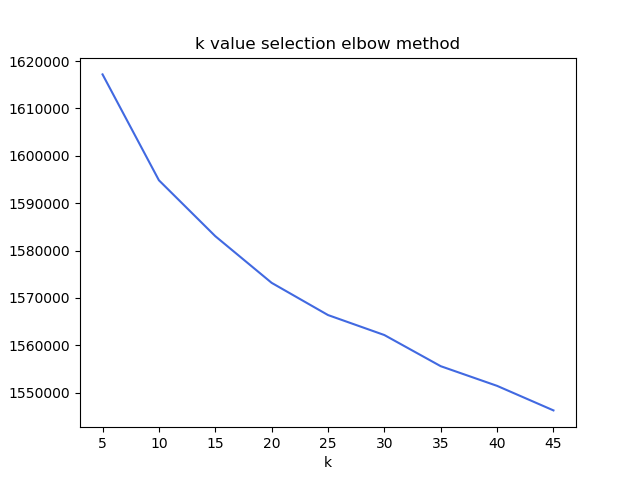}
    % \caption{SymbolOf\_EM}
  \end{minipage}
  \begin{minipage}[t]{0.31\textwidth}
    \centering
    \includegraphics[width=1.7in, height=2.0in]{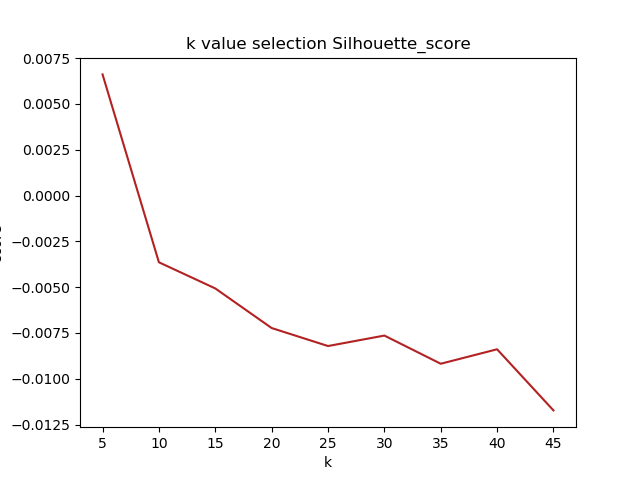}
    % \caption{SymbolOf\_Sil}
  \end{minipage}
  \begin{minipage}[t]{0.31\textwidth}
    \centering
    \includegraphics[width=1.7in, height=2.0in]{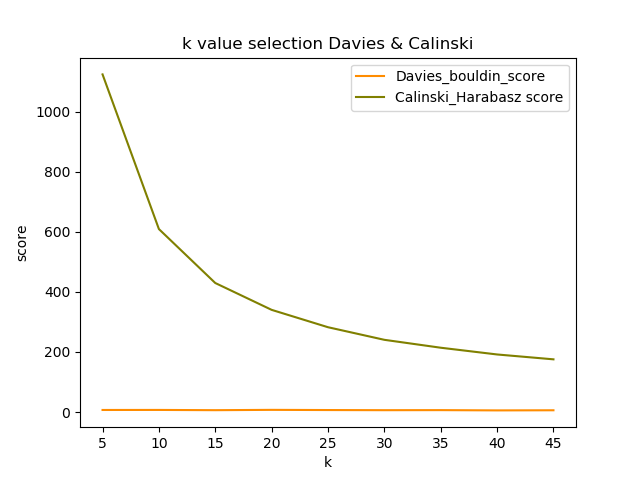}
    % \caption{SymbolOf\_DC}
  \end{minipage}
  \caption{\csentence{Selection of the parameter k for k-means clustering of SymbolOf triples.} The line chart describes the trend in the SymbolOf triples' clustering scores with different values of k. The clustering score used in the subplots (from left to right), is measured by the elbow method, Silhouette Coefficient, Davies-Bouldin index, and Calinski-Harabasz index, respectively. The x-axis shows the k value and the score is shown on the y-axis.}
    \label{figure:symbolof-para}
\end{figure} 

\begin{figure}[htbp]
  \begin{minipage}[t]{0.31\textwidth}
    \centering
    \includegraphics[width=1.7in, height=2.0in]{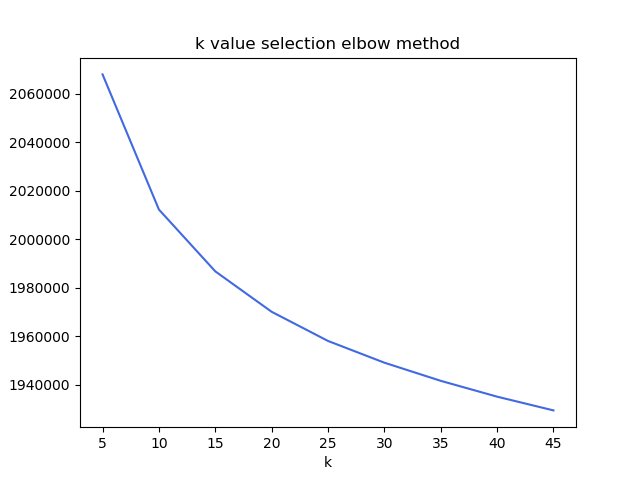}
  \end{minipage}
  \begin{minipage}[t]{0.31\textwidth}
    \centering
    \includegraphics[width=1.7in, height=2.0in]{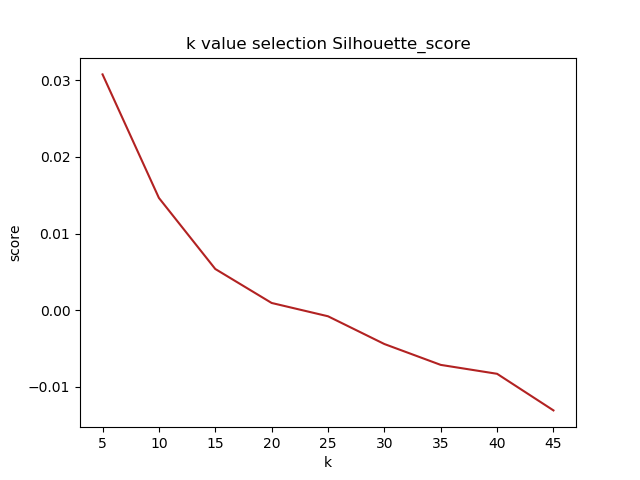}
  \end{minipage}
  \begin{minipage}[t]{0.31\textwidth}
    \centering
    \includegraphics[width=1.7in, height=2.0in]{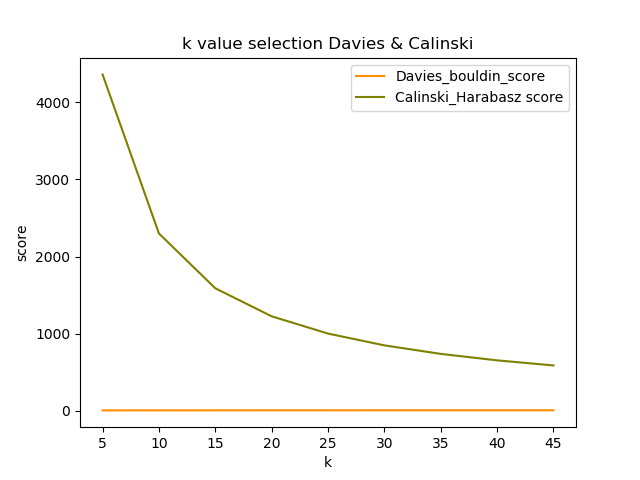}
  \end{minipage}
  \caption{\csentence{Selection of the parameter k for k-means clustering of FormOf triples.} The line chart describes the trend in the FormOf triples' clustering scores with different values of k. The clustering score used in the subplots (from left to right), is measured by the elbow method, Silhouette Coefficient, Davies-Bouldin index, and Calinski-Harabasz index, respectively. The x-axis shows the k value and the score is shown on the y-axis.}
    \label{figure:formof-para}
\end{figure} 

\begin{center}
\begin{longtable}[h!] {|c|c|c|}
\hline
\multicolumn{1}{|c|}{{ cluster ID}}         & \multicolumn{1}{c|}{{ head}}                 & \multicolumn{1}{c|}{{ tail}}              \\ \hline
\endfirsthead

\multicolumn{3}{c}%
{{\bfseries \tablename\ \thetable{} -- continued from previous page}} \\
\hline
\multicolumn{1}{|c|}{{ cluster ID}}         & \multicolumn{1}{c|}{{ head}}                 & \multicolumn{1}{c|}{{ tail}}              \\ \hline
\endhead

\hline \multicolumn{3}{|r|}{{Continued on next page}} \\ \hline
\endfoot

\endlastfoot
\multirow{5}{*}{0}                & /c/en/chifferobe/n              & /c/en/chifforobe            \\ \cline{2-3} 
                                  & /c/en/fluoroscans/n             & /c/en/fluoroscan            \\ \cline{2-3} 
                                  & /c/en/woodifying/v              & /c/en/woodify               \\ \cline{2-3} 
                                  & /c/pt/nintendinhos/n            & /c/pt/nintendinho           \\ \cline{2-3} 
                                  & /c/lb/maach/vn                   & /c/lb/maachen                 \\ \hline
\multirow{5}{*}{1}                & /c/es/ondearíais/v              & /c/es/ondear                \\ \cline{2-3} 
                                  & /c/it/istallerà/v               & /c/it/istallare             \\ \cline{2-3} 
                                  & /c/la/dextrarum/a               & /c/la/dexter                \\ \cline{2-3} 
                                  & /c/es/envasáramos/v             & /c/es/envasar               \\ \cline{2-3} 
                                  & /c/de/gestatte/v                & /c/de/gestatten             \\ \hline
\multirow{5}{*}{2}                & /c/la/redeatis/v                & /c/la/redeo                 \\ \cline{2-3} 
                                  & /c/la/ruminemur/v               & /c/la/ruminor               \\ \cline{2-3} 
                                  & /c/la/protrusi/v                & /c/la/protrudo              \\ \cline{2-3} 
                                  & /c/la/normabere/v         & /c/la/normo                   \\ \cline{2-3} 
                                  & /c/es/campean/v                 & /c/es/campear               \\ \hline
\multirow{5}{*}{3}                & /c/hu/hátsók/n                  & /c/hu/hátsó                 \\ \cline{2-3} 
                                  & /c/la/perspeculati              & /c/la/perspeculatus         \\ \cline{2-3} 
                                  & /c/de/iq tests                 & /c/de/iq\_test/n                   \\ \cline{2-3} 
                                  & /c/en/duck\_dives/n             & /c/en/duck\_dive            \\ \cline{2-3} 
                                  & /c/fr/argenteurs/n              & /c/fr/argenteur             \\ \hline
\multirow{5}{*}{4}                & /c/la/insutasv              & /c/la/insutus          \\ \cline{2-3} 
                                  & /c/fr/icosanoïde                & /c/fr/eicosanoïde/n         \\ \cline{2-3} 
                                  & /c/it/parlamenteremo/v          & /c/it/parlamentare          \\ \cline{2-3} 
                                  & /c/es/galardonase/v             & /c/es/galardonar            \\ \cline{2-3} 
                                  & /c/it/metterti/v                & /c/it/mettersi              \\ \hline
\multirow{5}{*}{5}                & /c/la/renidebam/v               & /c/la/renideo               \\ \cline{2-3} 
                                  & /c/it/cogestivano/v             & /c/it/cogestire             \\ \cline{2-3} 
                                  & /c/la/configurabimus/v          & /c/la/configuro             \\ \cline{2-3} 
                                  & /c/it/disincanterebbero/v       & /c/it/disincantare          \\ \cline{2-3} 
                                  & /c/la/fundassent/v              & /c/la/fundo                 \\ \hline
\multirow{5}{*}{6}                & /c/es/ventilé/v                 & /c/es/ventilar              \\ \cline{2-3} 
                                  & /c/es/embargases/v              & /c/es/embargar              \\ \cline{2-3} 
                                  & /c/es/vinculando/v              & /c/es/vincular              \\ \cline{2-3} 
                                  & /c/fr/cylindrasses/v            & /c/fr/cylindrer             \\ \cline{2-3} 
                                  & /c/fr/classai/v                 & /c/fr/classer               \\ \hline
\multirow{5}{*}{7}                & /c/it/ratificasti/v             & /c/it/ratificare            \\ \cline{2-3} 
                                  & /c/es/pichicateé/v              & /c/es/pichicatear           \\ \cline{2-3} 
                                  & /c/es/blandías/v                & /c/es/blandir               \\ \cline{2-3} 
                                  & /c/it/girandoti/v               & /c/it/girarsi               \\ \cline{2-3} 
                                  & /c/it/sabbierò/v                & /c/it/sabbiare              \\ \hline
\multirow{5}{*}{8}                & /c/la/insultasset/v             & /c/la/insulto               \\ \cline{2-3} 
                                  & /c/la/concutitote/v             & /c/la/concutio              \\ \cline{2-3} 
                                  & /c/la/adprobaremus/v            & /c/la/adprobo               \\ \cline{2-3} 
                                  & /c/la/consepsissemus/v          & /c/la/consepio              \\ \cline{2-3} 
                                  & /c/la/variegarentur/v           & /c/la/variego               \\ \hline
\multirow{5}{*}{9}                & /c/de/liebhabers                & /c/de/liebhaber/n           \\ \cline{2-3} 
                                  & /c/de/kalbst                    & /c/de/kalben/v              \\ \cline{2-3} 
                                  & /c/de/emblemen                  & /c/de/emblem/n              \\ \cline{2-3} 
                                  & /c/de/unseres/n                 & /c/de/unser                 \\ \cline{2-3} 
                                  & /c/de/absurdem/a                & /c/de/absurd                \\ \hline
\multirow{5}{*}{10}               & /c/es/retasteis/v               & /c/es/retar                 \\ \cline{2-3} 
                                  & /c/es/canalizarlas/v            & /c/es/canalizar             \\ \cline{2-3} 
                                  & /c/es/abombes/v                 & /c/es/abombar               \\ \cline{2-3} 
                                  & /c/es/transportada/v            & /c/es/transportar           \\ \cline{2-3} 
                                  & /c/fr/brancarderait/v           & /c/fr/brancarder            \\ \hline
\multirow{5}{*}{11}               & /c/nrf/boulets\_contraceptifs/n & /c/nrf/boulet\_contraceptif \\ \cline{2-3} 
                                  & /c/de/schwammerls/n              & /c/de/schwammerl/n             \\ \cline{2-3} 
                                  & /c/is/álfu                      & /c/is/álfa/n                \\ \cline{2-3} 
                                  & /c/es/mosqueándola/v            & /c/es/mosqueando            \\ \cline{2-3} 
                                  & /c/it/disgregandoti/v           & /c/it/disgregarsi           \\ \hline
\multirow{5}{*}{12}               & /c/en/pennames/n                & /c/en/penname               \\ \cline{2-3} 
                                  & /c/en/falcades                  & /c/en/falcade/n             \\ \cline{2-3} 
                                  & /c/en/electrical\_engineers     & /c/en/electrical\_engineer  \\ \cline{2-3} 
                                  & /c/ku/transitiv                 & /c/ku/berdan/v              \\ \cline{2-3} 
                                  & /c/sv/klassrums/n               & /c/sv/klassrum              \\ \hline
\multirow{5}{*}{13}               & /c/la/propinaremur/v            & /c/la/propino               \\ \cline{2-3} 
                                  & /c/la/friarat/v                 & /c/la/frio                  \\ \cline{2-3} 
                                  & /c/la/lucubrere/v               & /c/la/lucubro               \\ \cline{2-3} 
                                  & /c/la/labascimur/v              & /c/la/labasco               \\ \cline{2-3} 
                                  & /c/la/interimenti               & /c/la/interimens            \\ \hline
\multirow{5}{*}{14}               & /c/es/progresás/v               & /c/es/progresar             \\ \cline{2-3} 
                                  & /c/es/tintáremos/v              & /c/es/tintar                \\ \cline{2-3} 
                                  & /c/es/resentíamos/v             & /c/es/resentirse            \\ \cline{2-3} 
                                  & /c/la/centesimabere/v           & /c/la/centesimo             \\ \cline{2-3} 
                                  & /c/la/perviguerunt/v            & /c/la/pervigeo              \\ \hline
\multirow{5}{*}{15}               & /c/fr/entradmirera/v            & /c/fr/entradmirer           \\ \cline{2-3} 
                                  & /c/es/coagularé/v               & /c/es/coagular              \\ \cline{2-3} 
                                  & /c/frp/itèila                   & /c/frp/esèila/n             \\ \cline{2-3} 
                                  & /c/la/furcillas/n               & /c/la/furcilla              \\ \cline{2-3} 
                                  & /c/en/emanations/n              & /c/en/emanation             \\ \hline
\multirow{5}{*}{16}               & /c/no/dådyret/n                 & /c/no/dådyr                 \\ \cline{2-3} 
                                  & /c/nl/oppakt/v                  & /c/nl/oppakken              \\ \cline{2-3} 
                                  & /c/de/anwaltsgehilfin           & /c/de/anwaltsgehilfe/n      \\ \cline{2-3} 
                                  & /c/fi/kaverisektoriin/n         & /c/fi/kaverisektori         \\ \cline{2-3} 
                                  & /c/it/chicchiriai/v                & /c/it/chicchiriare             \\ \hline
\multirow{5}{*}{17}               & /c/it/miracoleremmo/v               & /c/it/miracolare       \\ \cline{2-3} 
                                  & /c/la/inaudias/v                & /c/la/inaudio            \\ \cline{2-3} 
                                  & /c/la/extergamus/v           & /c/la/extergo          \\ \cline{2-3} 
                                  & /c/la/allegavisse/v              & /c/la/allego            \\ \cline{2-3} 
                                  & /c/de/anarchischeres/a          & /c/de/anarchisch            \\ \hline
\multirow{5}{*}{18}               & /c/fr/remailles/v               & /c/fr/remailler             \\ \cline{2-3} 
                                  & /c/en/ebbtide/n                 & /c/en/ebb\_tide             \\ \cline{2-3} 
                                  & /c/fr/froufrouterait/v          & /c/fr/froufrouter           \\ \cline{2-3} 
                                  & /c/fr/attireront/v              & /c/fr/attirer               \\ \cline{2-3} 
                                  & /c/en/languaging/v              & /c/en/language              \\ \hline
\multirow{5}{*}{19}               & /c/sv/folkminnens/n             & /c/sv/folkminne             \\ \cline{2-3} 
                                  & /c/la/eluctari/v                & /c/la/eluctor               \\ \cline{2-3} 
                                  & /c/la/adpares/v                 & /c/la/adparo                \\ \cline{2-3} 
                                  & /c/es/calcinarme/v              & /c/es/calcinar              \\ \cline{2-3} 
                                  & /c/pt/alocaríeis/v              & /c/pt/alocar                \\ \hline

\caption{Example triples in 20 FormOf subclusters. The cluster IDs are consistent with those used in Figure \ref{figure: FormOf}.}
\label{table: FormOf}
\end{longtable}
\end{center}

\begin{center}
\begin{longtable}[h!] {|c|c|c|}
\hline

\multicolumn{1}{|c|}{{ cluster ID}}         & \multicolumn{1}{c|}{{ head}}                 & \multicolumn{1}{c|}{{ tail}}              \\ \hline
\endfirsthead

\multicolumn{3}{l}%
{{\bfseries \tablename\ \thetable{} -- continued from previous page}} \\
\hline
\multicolumn{1}{|c|}{{ cluster ID}}         & \multicolumn{1}{c|}{{ head}}                 & \multicolumn{1}{c|}{{ tail}}              \\ \hline
\endhead

\hline \multicolumn{3}{|r|}{{Continued on next page}} \\ \hline
\endfoot

\endlastfoot
\multirow{5}{*}{0}  & /c/en/immunosenescent/a   & /c/en/pathology                       \\ \cline{2-3} 
                    & /c/en/handball/v          & /c/en/soccer                          \\ \cline{2-3} 
                    & /c/en/screenplay/n        & /c/fr/cinéma                          \\ \cline{2-3} 
                    & /c/en/crivvens            & /c/en/scotland                        \\ \cline{2-3} 
                    & /c/en/rhizomatic/a        & /c/en/philosophy                      \\ \hline
\multirow{5}{*}{1}  & /c/fr/sapide/a        & /c/en/literary                        \\ \cline{2-3} 
                    & /c/hu/szirn/n               & /c/en/literary                        \\ \cline{2-3} 
                    & /c/ga/eo/n/wikt/en\_3     & /c/en/literary                        \\ \cline{2-3} 
                    & /c/af/elk/n               & /c/en/literary                        \\ \cline{2-3} 
                    & /c/ga/gair/v/wikt/en\_1   & /c/en/literary                        \\ \hline
\multirow{5}{*}{2}  & /c/it/vena\_cava/n        & /c/en/anatomy                         \\ \cline{2-3} 
                    & /c/et/fluor/n             & /c/fr/chimie                          \\ \cline{2-3} 
                    & /c/fr/saksaoul/n          & /c/fr/botanique                       \\ \cline{2-3} 
                    & /c/mul/raw                & /c/fr/linguistique                    \\ \cline{2-3} 
                    & /c/fr/tagbanoua/n         & /c/fr/linguistique                    \\ \hline
\multirow{5}{*}{3}  & /c/fr/schipluiden/n       & /c/fr/localités                       \\ \cline{2-3} 
                    & /c/en/brontotherid/n      & /c/en/zoology                         \\ \cline{2-3} 
                    & /c/fr/de\_hem/n           & /c/fr/localités                       \\ \cline{2-3} 
                    & /c/fr/brozolo/n           & /c/fr/localités                       \\ \cline{2-3} 
                    & /c/en/onychoteuthid/n     & /c/en/zoology                         \\ \hline
\multirow{5}{*}{4}  & /c/ha/umra/n              & /c/en/islam                           \\ \cline{2-3} 
                    & /c/no/jordakse/n          & /c/en/geometry                        \\ \cline{2-3} 
                    & /c/en/tsar/n              & /c/en/historical                      \\ \cline{2-3} 
                    & /c/lij/dîsètte            & /c/en/cardinal                        \\ \cline{2-3} 
                    & /c/mi/iwa                 & /c/en/cardinal                        \\ \hline
\multirow{5}{*}{5}  & /c/fr/scheelite/n         & /c/en/mineral                         \\ \cline{2-3} 
                    & /c/de/natriumdichromat/n  & /c/en/inorganic\_compound             \\ \cline{2-3} 
                    & /c/en/oxazepane/n         & /c/en/organic\_compound               \\ \cline{2-3} 
                    & /c/en/gelsemine/n         & /c/en/organic\_compound               \\ \cline{2-3} 
                    & /c/en/conductin/n         & /c/en/protein                         \\ \hline
\multirow{5}{*}{6}  & /c/en/azodicarbonamide/n  & /c/en/chemistry                       \\ \cline{2-3} 
                    & /c/en/ricinoleate/n       & /c/en/chemistry                       \\ \cline{2-3} 
                    & /c/fi/rikkiyhdiste/n      & /c/en/chemistry                       \\ \cline{2-3} 
                    & /c/en/test/v/wikt/en\_1   & /c/en/chemistry                       \\ \cline{2-3} 
                    & /c/en/vinyl\_acetate/n    & /c/en/chemistry                       \\ \hline
\multirow{5}{*}{7}  & /c/en/business/n          & /c/en/los\_angeles                    \\ \cline{2-3} 
                    & /c/sq/shkretëroj/v        & /c/en/tosk                            \\ \cline{2-3} 
                    & /c/en/hooklet/n           & /c/en/natural\_history                \\ \cline{2-3} 
                    & /c/da/femten                 & /c/en/cardinal            \\ \cline{2-3} 
                    & /c/it/un/a                & /c/en/sometimes\_before\_vowel\_or\_h \\ \hline
\multirow{5}{*}{8}  & /c/fr/déontologie/n       & /c/en/philosophy                      \\ \cline{2-3} 
                    & /c/en/cap\_cloud/n        & /c/en/meteorology                     \\ \cline{2-3} 
                    & /c/en/back\_ganging/n     & /c/en/uk                              \\ \cline{2-3} 
                    & /c/en/syringic/a          & /c/en/organic\_chemistry              \\ \cline{2-3} 
                    & /c/en/diethenoid/a        & /c/en/organic\_chemistry              \\ \hline
\multirow{5}{*}{9}  & /c/en/meteor/n            & /c/en/martial\_arts                   \\ \cline{2-3} 
                    & /c/fr/pause/n             & /c/fr/musique                         \\ \cline{2-3} 
                    & /c/cs/moderátor/n         & /c/en/uk                              \\ \cline{2-3} 
                    & /c/en/lin/v/wikt/en\_1    & /c/en/uk                              \\ \cline{2-3} 
                    & /c/ms/kata\_benda/n       & /c/en/grammar                         \\ \hline
\multirow{5}{*}{10} & /c/en/neurodegeneration/n & /c/fr/neurologie                      \\ \cline{2-3} 
                    & /c/no/oppholde/v          & /c/en/somewhere                       \\ \cline{2-3} 
                    & /c/scn/lu                 & /c/en/definite\_article               \\ \cline{2-3} 
                    & /c/en/monotypy/n          & /c/en/conservation\_biology           \\ \cline{2-3} 
                    & /c/nl/wao/n      & /c/en/netherlands                        \\ \hline
\multirow{5}{*}{11} & /c/sl/kriptozoologija/n   & /c/fr/biologie                        \\ \cline{2-3} 
                    & /c/nl/neptunus/n          & /c/fr/astronomie                      \\ \cline{2-3} 
                    & /c/fr/l1/n/wikt/fr\_2     & /c/fr/astronomie                      \\ \cline{2-3} 
                    & /c/en/freedom\_rider/n    & /c/en/politics                        \\ \cline{2-3} 
                    & /c/fr/corps/n             & /c/fr/numismatique                    \\ \hline
\multirow{5}{*}{12} & /c/en/vamptastic/a        & /c/en/slang                           \\ \cline{2-3} 
                    & /c/de/funzen/v/wikt/en\_1 & /c/en/slang                           \\ \cline{2-3} 
                    & /c/fi/filu/n              & /c/en/computing                       \\ \cline{2-3} 
                    & /c/en/dep/n               & /c/en/computing                       \\ \cline{2-3} 
                    & /c/en/nonserver/a         & /c/en/computing                       \\ \hline
\multirow{5}{*}{13} & /c/en/fiscal/n/wikt/en\_1 & /c/en/legal                           \\ \cline{2-3} 
                    & /c/de/silver\_goal/n      & /c/en/football                        \\ \cline{2-3} 
                    & /c/en/shitcan/v           & /c/en/vulgar                          \\ \cline{2-3} 
                    & /c/en/eicosanoid/n        & /c/fr/biochimie                       \\ \cline{2-3} 
                    & /c/rm/mel/n               & /c/en/rumantsch\_grischun             \\ \hline
\multirow{5}{*}{14} & /c/fr/océan\_atlantique/n & /c/fr/géographie                      \\ \cline{2-3} 
                    & /c/sl/balkanski/a         & /c/fr/géographie                      \\ \cline{2-3} 
                    & /c/fr/riviera/n           & /c/fr/géographie                      \\ \cline{2-3} 
                    & /c/alt/cyy/n              & /c/fr/géographie                      \\ \cline{2-3} 
                    & /c/fr/sapouy/n            & /c/fr/géographie                      \\ \hline
\multirow{5}{*}{15} & /c/fr/lapedona/n          & /c/fr/localités                       \\ \cline{2-3} 
                    & /c/fr/tour\_de\_faure/n   & /c/fr/localités                       \\ \cline{2-3} 
                    & /c/fr/amendeuix\_oneix/n  & /c/fr/localités                       \\ \cline{2-3} 
                    & /c/fr/espédaillac/n       & /c/fr/localités                       \\ \cline{2-3} 
                    & /c/fr/rye/n               & /c/fr/localités                       \\ \hline
\multirow{5}{*}{16} & /c/fro/voleir/n           & /c/en/anglo\_norman                   \\ \cline{2-3} 
                    & /c/en/stoater/n           & /c/en/horse\_racing                   \\ \cline{2-3} 
                    & /c/nrf/malon/n            & /c/en/jersey                          \\ \cline{2-3} 
                    & /c/en/antieczema/a        & /c/en/pharmacology                    \\ \cline{2-3} 
                    & /c/ga/heitribhéascna/n    & /c/en/linguistics                     \\ \hline
\multirow{5}{*}{17} & /c/fr/humoral/a           & /c/fr/médecine                        \\ \cline{2-3} 
                    & /c/en/fasciculatory/a     & /c/en/medicine                        \\ \cline{2-3} 
                    & /c/fr/antiépileptique/a   & /c/fr/médecine                        \\ \cline{2-3} 
                    & /c/de/tropf/n             & /c/fr/médecine                        \\ \cline{2-3} 
                    & /c/fr/sida/n/wikt/fr\_1   & /c/fr/médecine                        \\ \hline
\multirow{5}{*}{18} & /c/en/polymodality/n      & /c/en/mathematics                     \\ \cline{2-3} 
                    & /c/en/biplanar/a          & /c/en/mathematics                     \\ \cline{2-3} 
                    & /c/it/esaedro/n           & /c/en/mathematics                     \\ \cline{2-3} 
                    & /c/de/divergieren/v       & /c/en/mathematics                     \\ \cline{2-3} 
                    & /c/en/local\_maximum/n    & /c/en/mathematics                     \\ \hline
\multirow{5}{*}{19} & /c/en/thermoelasticity/n  & /c/en/physics                         \\ \cline{2-3} 
                    & /c/pt/hidrostático/a      & /c/en/physics                         \\ \cline{2-3} 
                    & /c/en/remanence/n         & /c/en/physics                         \\ \cline{2-3} 
                    & /c/en/specific/a          & /c/en/physics                         \\ \cline{2-3} 
                    & /c/en/microelectronvolt/n & /c/en/physics                         \\ \hline
\caption{Example triples in 20 HasContext subclusters. The cluster IDs are consistent with those used in Figure \ref{figure: hascontext}.}
\label{table: HasContext}
\end{longtable}
\end{center}

\begin{figure}[htbp]
  \begin{minipage}[t]{0.31\textwidth}
    \centering
    \includegraphics[width=1.59in]{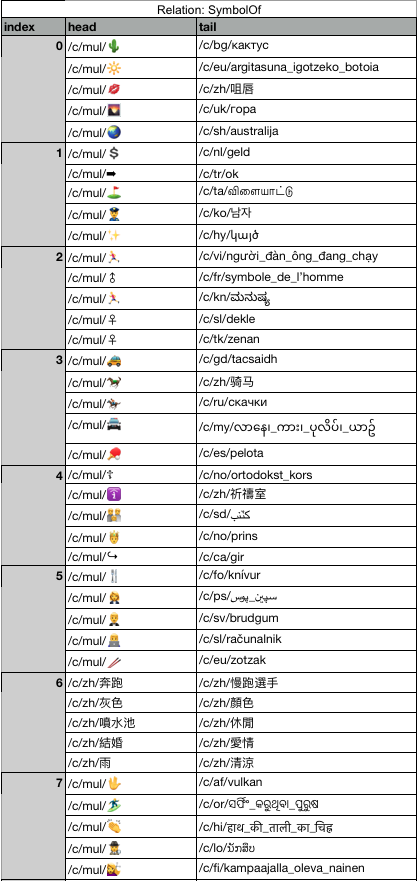}
  \end{minipage}
  \begin{minipage}[t]{0.31\textwidth}
    \centering
    \includegraphics[width=1.59in]{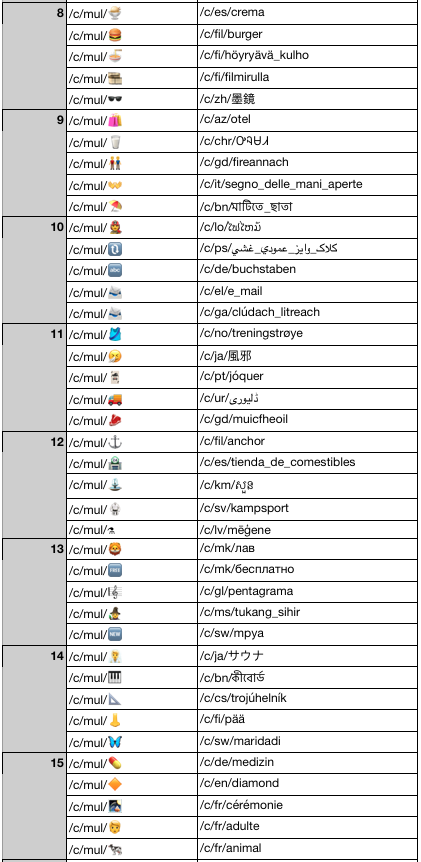}
  \end{minipage}
  \begin{minipage}[t]{0.31\textwidth}
    \centering
    \includegraphics[width=1.59in]{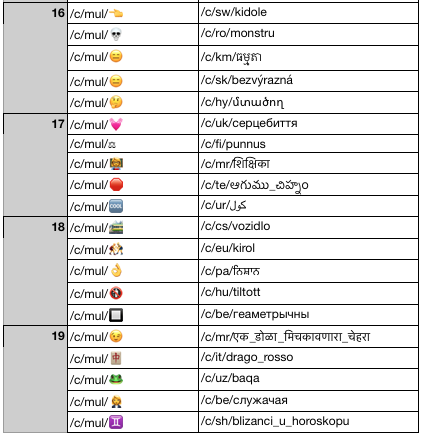}
  \end{minipage}
  \caption{\csentence{Example triples in SymbolOf's 20 subclusters} Because most entities in SymbolOf triples are emojis, we take a screenshot of triples-examples in 20 SymbolOf  clusters rather than show these examples in a formal table. For each cluster, we randomly select 5 examples; the index in the first column indicates the triple's cluster ID. These cluster IDs are consistent with those in Figure \ref{figure: SymbolOf}, Table \ref{table: average_cohesion}, and Table \ref{table: average_separation}}.
  \label{figure: SymbolOf_example}
\end{figure}

% if your bibliography is in bibtex format, use those commands:
\bibliographystyle{bmc-mathphys} % Style BST file (bmc-mathphys, vancouver, spbasic).
\bibliography{Refernces}      % Bibliography file (usually '*.bib' )

\end{backmatter}
\end{document}